\documentclass[10pt]{article} 

\usepackage{xcolor}
\definecolor{myblue}{RGB}{30, 100, 200}

\usepackage{graphicx}
\usepackage{subcaption}
\usepackage{booktabs}
\usepackage{multirow}
\usepackage{mdframed}
\usepackage[table]{xcolor}
\definecolor{mygray}{RGB}{220,220,220} 
\usepackage[accepted]{tmlr}

\usepackage{amsmath,amsfonts,bm}

\def\eqref#1{equation~\ref{#1}}

\def\1{\bm{1}}

\DeclareMathAlphabet{\mathsfit}{\encodingdefault}{\sfdefault}{m}{sl}
\SetMathAlphabet{\mathsfit}{bold}{\encodingdefault}{\sfdefault}{bx}{n}

\usepackage{cancel}
\usepackage{hyperref}
\usepackage{url}

\title{ActAlign: Zero-Shot Fine-Grained Video Classification \\
via Language-Guided Sequence Alignment}

\author{\name Amir Aghdam \email amir.aghdam@temple.edu \\
      \addr Department of Computer Science, Temple University\\
      Philadelphia, PA, USA
      \AND
      \name Vincent Tao Hu \email taohu620@gmail.com \\
      \addr CompVis @ LMU Munich, Munich Center for Machine Learning \\
      Munich, Germany
      \AND
      \name Bjorn Ommer \email bommer@lmu.de\\
      \addr CompVis @ LMU Munich, Munich Center for Machine Learning \\
      Munich, Germany}

\def\month{10}  
\def\year{2025} 
\def\openreview{\url{https://openreview.net/forum?id=Nwzn4qMTGb}} 

\begin{document}

\maketitle

\begin{abstract}
We address the task of zero-shot video classification for extremely fine-grained actions (e.g., Windmill Dunk in basketball), where no video examples or temporal annotations are available for unseen classes. While image–language models (e.g., CLIP, SigLIP) show strong open-set recognition, they lack temporal modeling needed for video understanding. We propose ActAlign, a \textit{truly zero-shot}, training-free method that formulates video classification as a sequence alignment problem, preserving the generalization strength of pretrained image–language models. For each class, a large language model (LLM) generates an ordered sequence of sub-actions, which we align with video frames using Dynamic Time Warping (DTW) in a shared embedding space. Without any video–text supervision or fine-tuning, ActAlign achieves 30.4\% accuracy on ActionAtlas—the most diverse benchmark of fine-grained actions across multiple sports—where human performance is only 61.6\%. ActAlign outperforms billion-parameter video–language models while using \textbf{$\sim$8$\times$ fewer parameters}. Our approach is \textit{model-agnostic and domain-general}, demonstrating that structured language priors combined with classical alignment methods can unlock the open-set recognition potential of image–language models for fine-grained video understanding. \\
\textbf{\textit{Code Link}}: \textit{\href{https://amir-aghdam.github.io/act-align/}{https://amir-aghdam.github.io/act-align/}}
\end{abstract}


\section{Introduction}
\begin{figure}[t]
    \centering
    {\small \textbf{Image-Text Embedding Space}} \\[0.2em]
    \includegraphics[width=0.6\linewidth]{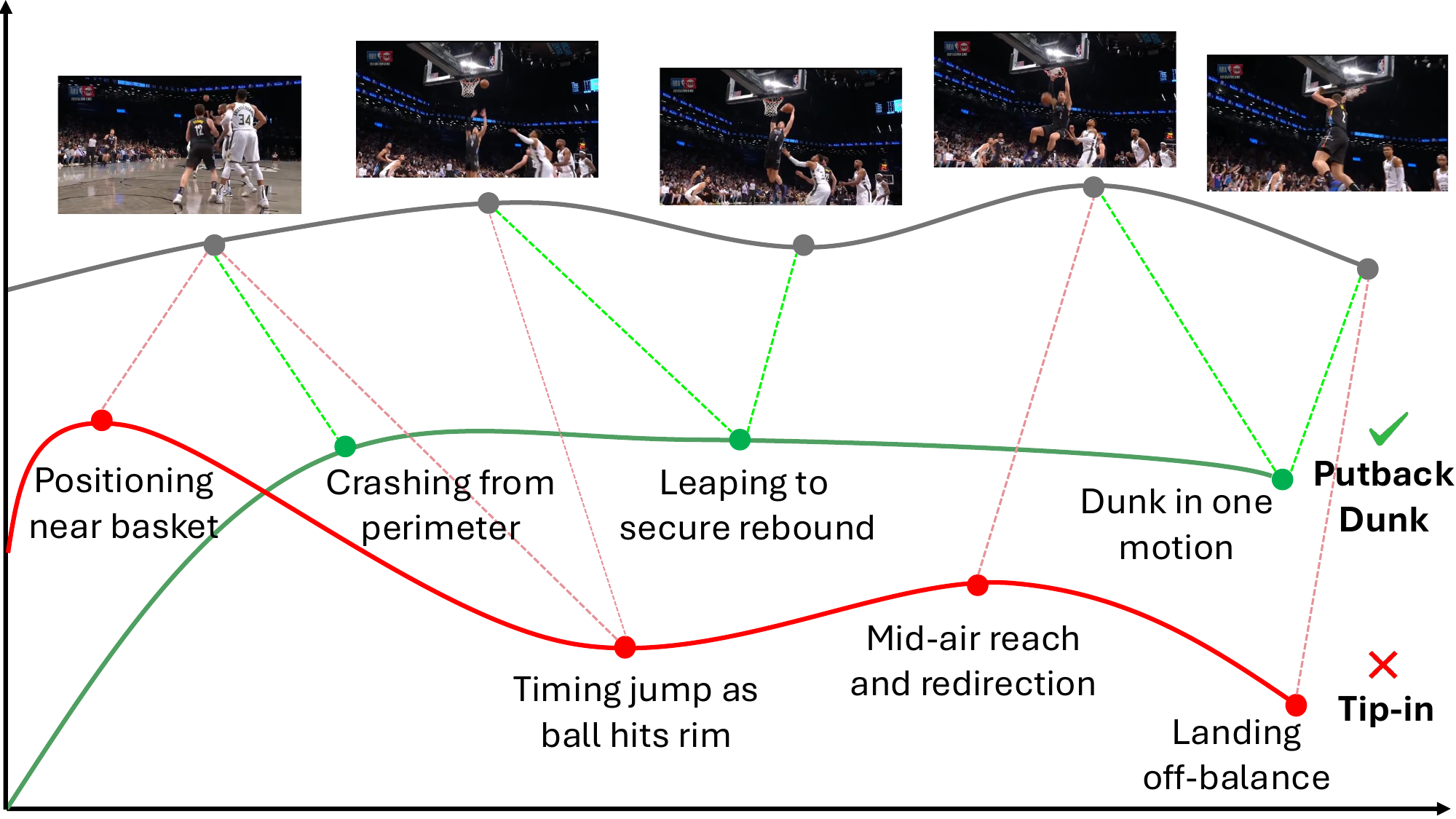}
    \caption{{ActAlign improves zero-shot fine-grained action recognition by modeling them as structured language sequences. By aligning sub-action descriptions with video frames (green vs. red paths), we achieve more accurate predictions \textbf{without requiring any video-text training data.}}
    }
    \label{fig:teaser}
\end{figure}

Understanding fine-grained human activities in video—such as distinguishing a \emph{hook shot} from a \emph{layup} in basketball, or recognizing tactical formations in football—requires parsing subtle, temporally ordered visual cues across frames. These fine-grained actions unfold in structured sequences of sub-actions and are often nearly indistinguishable from one another in appearance. In contrast to general activities like swimming, which can often be inferred from a single frame showing a person in water, fine-grained recognition demands attention to temporally extended object interactions, spatial relations, and high-level intent. As such, models must not only understand what is present in a video but also \emph{when} and \emph{how} key sub-actions occur. This requires accurately aligning the temporal progression of sub-actions with each fine-grained action to ensure a correct prediction, as shown in Figure~\ref{fig:teaser}. While their high-level label is barely recognized by action recognition models (e.g. such as \emph{Roundhouse kick in Kickboxing}).

At the same time, contrastive vision--language models such as CLIP~\cite{radford2021clip} and SigLIP~\cite{Zhai2023SigmoidLF} have demonstrated impressive open-set fine-grained recognition in image domains by training on massive image--text pairs. These models learn a shared image--text latent space, which allows both visual inputs and natural language descriptions to be embedded for direct comparison without task-specific supervision. This enables zero-shot classification using natural language prompts and has been widely adopted for downstream recognition tasks. However, extending these capabilities to video understanding introduces new challenges and requires temporal modeling. Existing methods that adapt CLIP-style models to video recognition either average frame-level features~\cite{rasheed2023vifi, Zohra_2025_CVPR}—ignoring temporal structure—or fine-tune on target datasets~\cite{wang2022actionclip, ni2022xclip,10.1007/978-3-031-72664-4_5, Wang_Xing_Jiang_Chen_Mei_Zuo_Dai_Wang_Liu_2024}, sacrificing generalization and open-set recognition. In both cases, the fine-grained temporal semantics of actions are lost or diluted.

Recent video–language architectures and instruction-tuned LLM-based systems such as Video-LLaMA~\cite{zhang2023videollama}, VideoChat~\cite{li2023videochat}  mPLUG-Owl~\cite{ye2023mplug}, Qwen2-VL~\cite{wang2024qwen2vl}, and DeepSeek-JanusPro~\cite{chen2025janus} enable open-ended, dialog-style video understanding through heavy instruction tuning, but they are not tailored for fine-grained video recognition. 

Meanwhile, textual grounding (image--text alignment) remains a central challenge in interactive video-language models, especially for open-set and fine-grained video recognition. Dynamic Time Warping (DTW)~\cite{Vintsyuk1968}, a classical algorithm for aligning temporally mismatched sequences, has seen renewed interest through differentiable variants~\cite{dogan2018neuromatch, chang2019d3tw} designed for supervised image--text temporal alignment. Yet, these methods rely either on annotated transcripts or example support videos or originally proposed for supervised training, making them impractical for zero-shot recognition. Likewise, approaches using part-level or attribute-level supervision~\cite{wu2023bike, zhu2024purls} offer fine-grained cues but lack the ability to model the temporal structure between language-defined actions and visual content.

In this work, we introduce ActAlign, a novel framework that brings the open-set generalization power of image--text models to video classification of \textit{extremely fine-grained actions} through language-guided sub-action alignment in a truly \textit{zero-shot setting}. Rather than tuning a model for a specific domain or collapsing the video into a static representation, ActAlign operates in a training-free setting: for each unseen action class, we use a large language model (LLM) to generate a structured sequence of temporal sub-actions that semantically define the class. Then, using the pretrained SigLIP model~\cite{Zhai2023SigmoidLF} to extract frame-wise visual and sub-action features, we align the frame sequence with the LLM-generated sub-action script via Dynamic Time Warping (DTW) (see Figure~\ref{fig:teaser}). This allows us to compute a soft alignment score between different action classes that respects both content and temporal ordering, enabling fine-grained classification in a truly zero-shot manner.


\medskip
\textbf{Our contributions are as follows:}
\begin{itemize}
    \item We introduce a novel approach for zero-shot fine-grained video recognition that models each action as a \textit{general, structured temporal sequence of sub-actions} derived solely from action names---\textbf{without access to videos or transcripts}.
    
    \item We propose ActAlign, a \emph{domain-general, model-agnostic} and zero-shot framework that applies the open-set generalization strength of image--text models to the challenging task of fine-grained and zero-shot video classification by \textbf{reformulating the task to sequence matching} without requiring any video--text supervision.
    
    \item We demonstrate that ActAlign surpasses zero-shot and CLIP-based model baselines, outperforming even billion-parameter video–language models on the most challenging and diverse benchmark to our knowledge, ActionAtlas—where human-level accuracy caps at 61.64\%.
\end{itemize}




\section{Related Work}

\subsection{Action Recognition}
Action recognition has been extensively studied, with early work leveraging hand-crafted features and classic classifiers~\cite{1467360, jain2013better, 6751553}. The field has since evolved to deep learning approaches~\cite{carreira2017quo, 10.5555/2968826.2968890, Tran_2018_CVPR, 10.1007/978-3-319-46484-8_2}, including CNNs, RNNs, and more recently, Transformer-based models~\cite{Arnab_2021_ICCV, pmlr-v139-bertasius21a, Kim_Lee_Cho_Lee_Heo_2025, Yoshida_Shibata_Terao_Okatani_Sugiyama_2025}, which model spatio-temporal dynamics more effectively. Fine-grained action recognition remains challenging due to subtle inter-class variations over time. Recent methods tackle this under supervised~\cite{Leong2022CombinedCT, 10.1145/3503161.3548046}, semi-supervised~\cite{10.1007/978-3-031-20080-9_33, 10.1007/978-3-031-73242-3_22}, and few-shot~\cite{10.1145/3581783.3612221, Hong_2021_ICCV, 10.1145/3474085.3475216} settings to mitigate annotation costs. Yet, they remain reliant on some level of fine-grained labeling and are constrained to specific domains.

\subsection{Vision-Language Models}

\subsubsection{Image–Language Models}
Foundational image–language models such as CLIP~\cite{radford2021clip}, SigLIP~\cite{Zhai2023SigmoidLF}, and ALIGN~\cite{jia2021align} learn joint image–text embeddings from large-scale image–caption pairs. Such modeling enables strong open-set recognition without task-specific supervision. These models are widely adopted as pretrained backbones for downstream tasks, including visual question answering~\cite{albef2021,frozen2021, li2022blip, steiner2024paligemma2familyversatile}, image captioning and generation~\cite{clipcap2021, simvlm2021}, and few-/zero-shot classification~\cite{coop2021, Khattak_Naeem_Naseer_VanGool_Tombari_2025}. However, they lack temporal modeling, which limits their open-set recognition capability to video inputs.

\subsubsection{Video–Language Models}
Early efforts in video modeling focused on self-supervised pretraining. Models like ActBERT~\cite{Zhu_2020_CVPR} and VideoBERT~\cite{Sun_2019_ICCV} applied masked language modeling to video frames. This enabled better transfer to video–language tasks through finetuning on paired video–text data~\cite{miech2020milnce, lei2021clipbert, xu-etal-2021-vlm, 10.5555/3600270.3602875, Li_2022_CVPR}. Recent progress in LLM reasoning has driven their integration with visual encoders, enabling open-ended video understanding via user prompts. These models~\cite{ye2023mplug, li2023videochat, zhang2024video, wang2024qwen2vl}, such as \textit{Video-LLaMA}~\cite{zhang2023videollama}, combine pretrained visual backbones with chat-centric LLMs to generate spatio-temporal reasoning in conversational settings. While these models excel at open-ended question answering, they require extensive instruction tuning and are not optimized for fine-grained video recognition.

\subsection{Sequence Alignment}
Sequence alignment is a long-standing problem in fields such as biology and signal processing. A widely used technique for aligning temporally misaligned signals is Dynamic Time Warping (DTW)\cite{10.5555/3000850.3000887, 10.1145/347090.347153, 1211511}, which computes the optimal warping path between two ordered sequences. Early work extended DTW to video tasks\cite{huang2016ctc, richard2018action}. For instance, a seminal approach~\cite{bojanowski2015weakly} proposed aligning videos to ordered scripts by enforcing the temporal order of events. With the rise of deep learning, DTW has been adapted into differentiable forms~\cite{dogan2018neuromatch, chang2019d3tw}, enabling greater flexibility and improved learning capacity. A notable example is \textit{OTAM}~\cite{cao2020otam}, which leverages DTW for representation learning in few-shot video classification by aligning each video to its support set. However, these methods often assume access to ground-truth transcripts for each video or rely on task-specific supervision, such as paired video–text exemplars.

\medskip

\paragraph{Our Work}
In contrast to prior methods that discard temporal structure, rely on task-specific supervision, or require support data for alignment, ActAlign introduces a language-guided sequence alignment framework for zero-shot fine-grained video classification.


\section{Method}

\subsection{Problem Definition} Let $\mathcal{D}={(V_i, y_i)}_{i=1}^N$ denote a dataset of $N$ videos, where each video $V_i$ is associated with a ground-truth fine-grained class label $y_i$ drawn from a set of $M$ candidate classes $\mathcal{Y}={c_1,\dots,c_M}$. Each video $V_i$ consists of a sequence of $T_i$ frames as defined in Eq.~\ref{dataset}:

\begin{equation}
V_i = \{\mathbf{v}_i^t\}_{t=1}^{T_i}, \quad \mathbf{v}_i^t \in \mathbb{R}^{H\times W \times 3},
\label{dataset}
\end{equation}

where $H$ and $W$ denote the frame height and width. In our zero-shot setting, no video examples of the target classes $\mathcal{Y}$ are used for training or tuning; only high-level action names $c_j$ are provided. The goal is to construct a function $f: \mathcal{V} \times \mathcal{Y} \to \mathbb{R}$ that effectively maps the sequence of video frames into their correct action class $y$. The predicted class label $\hat{y}_i$ for a video $V_i$ is given by Eq.~\ref{pred}:
\begin{equation}
\hat{y}_i = \arg\max_{c_m \in \mathcal{Y}} f(V_i, c_m).
\label{pred}
\end{equation}

To leverage semantic priors from LLMs, we automatically decompose each class label $c_m$ into an ordered, variable-length sequence of $K_m$ textual sub-actions, as defined in Eq.~\ref{eq:subaction_sequence}:

\begin{equation}
S_m = [s_{m,1}, s_{m,2}, \dots, s_{m,k_m}],
\label{eq:subaction_sequence}
\end{equation}

where $s_{m,k}$ is a concise natural-language description of the $k$-th step in executing action $m$.

\subsection{Our Method}
\begin{figure*}
  \centering
  
  \includegraphics[width=\textwidth]{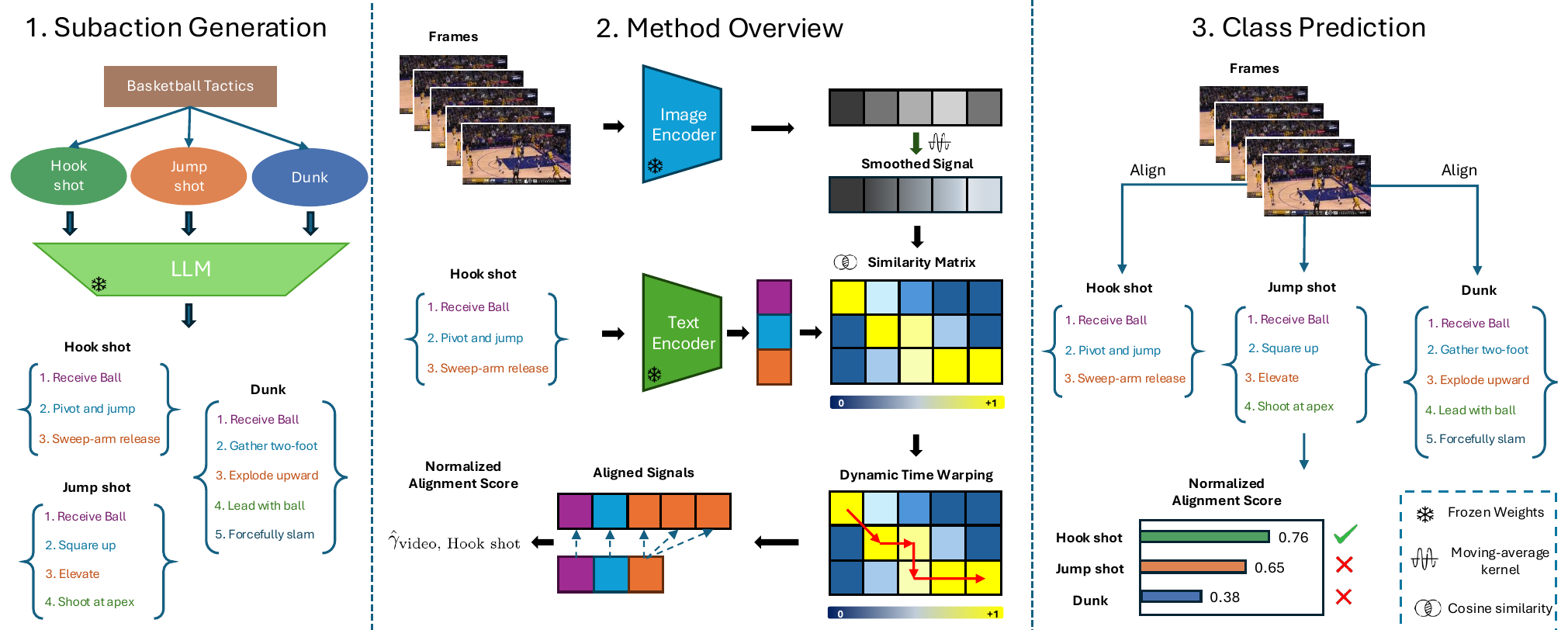}
  \caption{\textbf{Our ActAlign Method Overview.} (\textbf{1}) \emph{Sub-action Generation:} Given fine-grained actions (e.g.\ Basketball Tactics), we prompt an LLM to decompose each action (e.g.\ Hookshot, JumpShot, Dunk) into a temporal sequence of sub-actions. (\textbf{2}) \emph{Temporal Alignment:} Video frames are encoded by a frozen pretrained vision encoder and smoothed via a moving‐average filter. Simultaneously, each sub-action is encoded by the text encoder. We compute a cosine‐similarity matrix between frame and sub-action embeddings, then apply Dynamic Time Warping (DTW) to find the optimal alignment path and normalized alignment score. (\textbf{3}) \emph{Class Prediction:} We repeat this process for each candidate action \textit{m}, compare normalized alignment scores \(\hat{\gamma}_{\text{video}, m}\), and select the action sequence with the highest score as the final prediction. 
  } 
  \label{fig:pipeline}
\end{figure*}

We define $f(V_i, c_m)$ as the alignment score between the visual frame embeddings $\{\mathbf{v}_i^t\}$ and the sub-action sequence {$S_m$}, computed via Dynamic Time Warping (DTW). This alignment is performed in the image--text embedding space, without requiring any fine-tuning or video examples from the target label set. Figure~\ref{fig:pipeline} illustrates the pipeline of our proposed approach.

\subsection{Preliminary Sub-action Generation by LLM}
In domains requiring fine-grained distinctions—such as differentiating tactical plays in sports—the high-level action class name \( c_m \) often lacks sufficient discriminatory power. To address this, we define a mapping \( \mathcal{P}: \mathcal{Y} \to \mathcal{S} \), where \( \mathcal{S} \) denotes the space of ordered sub-action scripts. For each class \( c_m \in \mathcal{Y} \), we generate a sequence \( S_m = [s_{m,1}, \dots, s_{m,K_m}] = \mathcal{P}(c_m) \), where each \( s_{m,k} \) describes a trackable and temporally ordered sub-action.

We instantiate \( \mathcal{P} \) using GPT-4o via carefully constructed natural language prompts. Given a set of candidate classes \( \{c_1, \dots, c_M\} \), the LLM is instructed to:
\begin{enumerate}
    \item Decompose each \( c_m \) into a variable- or fixed-length sequence of semantically coherent sub-actions.
    \item Return the list \( [s_{m,1}, \dots, s_{m,K_m}] \) in a consistent, structured format.
\end{enumerate}

By leveraging the LLM’s extensive prior knowledge, we obtain sub-action sequences without any manual labels or video supervision. These sequences serve as semantic reference signals for temporal alignment.

\paragraph{Importance of Context in Sub-Actions.}  
Terse descriptions (e.g., “drive forward”) are often ambiguous and may refer to multiple sports, which hampers DTW alignment due to poor textual grounding. In contrast, context-rich prompts (e.g., “player drives forward to the rim”) produce more discriminative sub-actions that are reliably grounded with visual evidence.

\paragraph{Context Augmentation}
We enrich each sub-action by embedding it into a contextualized template:
\texttt{This is a video of doing <action name> in <sport domain> with <sub-action>}.
This substantially improves semantic clarity and further disambiguates sub-actions. The sport domain is automatically provided in the ActionAtlas metadata (and could also be inferred from the action name itself).

\subsection{Visual and Semantic Feature Encoding}
Once each class label $c_m$ is decomposed into its sub-action sequence $S_m$, we project both video frames and sub-actions into a shared $d$-dimensional embedding space using the pre-trained SigLIP image--text model, which is recognized for its strong zero-shot recognition performance.

\paragraph{Visual Embeddings} Let $\phi_v: \mathbb{R}^{H\times W\times 3} \to \mathbb{R}^d$ denote the vision encoder. For video $V_i = \{\mathbf{v}i^t\}_{t=1}^{T_i}$, we compute frame-level embeddings:
\begin{equation}
\mathbf{z}_i^t = \phi_v(\mathbf{v}_i^t) \in \mathbb{R}^d, \quad t=1,\dots, T_i.
\label{eq:frame_embeddings}
\end{equation}
Stacking yields $Z_i = [\mathbf{z}_i^1,\dots,\mathbf{z}_i^{T_i}] \in \mathbb{R}^{d\times T_i}$.

\paragraph{Semantic Embeddings}Let $\phi_t: \mathcal{T} \to \mathbb{R}^d$ be the  text encoder. For each class $c_m$ and its sub-action sequence $S_m$, we embed each step:
\begin{equation}
\mathbf{u}_{m,k} = \phi_t(s_{m,k}) \in \mathbb{R}^d, \quad k=1,\dots,K_m,
\label{eq:action_embedding}
\end{equation}
stitching into $U_m = [\mathbf{u}_{m,1},\dots,\mathbf{u}_{m,K_m}] \in \mathbb{R}^{d\times K_m}$.

\paragraph{Shared Latent Space} We normalize all embeddings so that the similarity
\begin{equation}
\text{sim}(\mathbf{z}_i^t,\mathbf{u}_{m, k}) = \frac{\mathbf{z}_i^t\top \mathbf{u}_{m, k}}{|\mathbf{z}_i^t|\cdot|\mathbf{u}_{m, k}|}
\label{eq:cosine_similarity}
\end{equation}
is a cosine similarity measure between frame $\mathbf{v}_i^t$ and sub-action $s_{m,k}$ of class $c_m$ (see Eq.~\ref{eq:cosine_similarity}). This cross-modality similarity forms the basis for alignment in the next step.
\subsection{Dynamic Time Warping}
After feature encoding, each video yields a visual embedding sequence $Z_i$ and ordered sequences of sub-action embeddings $U_m$ for each class $c_m$. We treat $U_m$ as the reference semantic signal and $\widetilde Z_i$ as the query visual signal. The $U_m$ could also be viewed as a prototype sequence for class $c_m$. 

\paragraph{Signal Smoothing}
Real-world footage often contains abrupt scene changes or irrelevant frames (e.g., replays, advertisements) that introduce noise into $Z_i$. To mitigate this, we apply a simple, parameter-free 1D moving-average filter of width $w$ across the temporal dimension:
\begin{equation}
\widetilde{\mathbf z}_i^t = \frac{1}{w} \sum_{\tau=t-\lfloor w/2\rfloor}^{t+\lfloor w/2\rfloor} \mathbf z_i^\tau,
\label{eq:signal_smoothing}
\end{equation}
with boundary conditions handled via zero padding. The kernel width $w$ controls the trade-off between noise reduction and temporal resolution. The smoothing effectively aggregates short-term temporal context, dampening isolated spikes while preserving the overall action dynamics. As it introduces no learnable parameters or additional training, it remains lightweight and fully compatible with the training-free setting.

\paragraph{Affinity Matrix Construction}

Let the smoothed visual embeddings for video \(V_i\) be
\(\widetilde Z_i = [\widetilde{\mathbf z}_i^1, \dots, \widetilde{\mathbf z}_i^{T_i}]\)
and the sub-action sequence embeddings for class \(c_m\) be
\(U_m = [\mathbf u_{m,1}, \dots, \mathbf u_{m,K_m}]\).
We first compute the raw cosine similarity matrix:

\begin{equation}
A^{(m,i)}_{k,t}
= \bigl\langle \mathbf{u}_{m,k},\, \widetilde{\mathbf{z}}_i^t \bigr\rangle,
\quad
A^{(m,i)} \in \mathbb{R}^{K_m \times T_i}
\label{eq:raw_similarity}
\end{equation}

where each \(\langle\cdot,\cdot\rangle\) is the inner product of $\mathcal{L}2$-normalized vectors, yielding values in \([-1,1]\).  
Following the SigLIP prediction approach, we then apply a \texttt{sigmoid} function \(\sigma(\cdot)\) to transform these values into affinity scores in \([0,1]\):

\begin{equation}
\hat A^{(m,i)}_{k,t}
= \sigma\bigl(\alpha\,A^{(m,i)}_{k,t} + \beta\bigr),
\label{eq:scaled_similarity}
\end{equation}

where \(\alpha,\beta\) are pre-learned scaling parameters as part of SigLIP's pretraining recipe. The resulting \(\hat A^{(i,m)}\) is used as the input affinity matrix for DTW alignment.

\paragraph{DTW Alignment and Scoring}

Given the affinity matrix $\hat{A}^{(m,i)}$ for class $c_m$ and video $V_i$ (defined in Eq.~\ref{eq:scaled_similarity}), we seek a warping path $W^{(m,i)} = \{(k_1,t_1),\dots,(k_L,t_L)\}$ that maximizes cumulative similarity under monotonicity and continuity constraints:

\begin{align}
W^{(m,i)} &= \arg\max_{W} \sum_{(k,t)\in W} \hat{A}^{(m,i)}_{k,t}, \nonumber \\
\text{s.t.} \quad & \textit{W is a valid warping path between } [1,K_m] \textit{ and } [1,T_i].
\label{eq:dtw_objective}
\end{align}

We solve this using dynamic programming by selecting the path with highest alignment (i.e. similarity value) at each step:

\begin{equation}
D_{k,t} = \hat{A}^{(m,i)}_{k,t} + \max \{D_{k,t-1}, D_{k-1,t}, D_{k-1,t-1}\},
\label{eq:dtw_recursion}
\end{equation}

with the base case $D_{0, *} = D_{*, 0} = -\infty$. The final alignment score is $\max_{k,t} D_{k,t}$, and backtracking recovers the optimal warping path $W^{(m,i)}$ for sub-action sequence of class $c_m$ and video $V_i$.

\paragraph{Prediction}
We compute the raw alignment score, as defined in Eq.~\ref{eq:raw_alignment}, as the sum of the similarity values of the optimal warping path $W^{(m,i)}$ obtained in Eq.~\ref{eq:dtw_objective}.

\begin{equation}
\gamma_{i,m} = \sum_{(k,t)\in W^{(m,i)}} \hat{A}^{(m,i)}_{k,t},
\label{eq:raw_alignment}
\end{equation}

$\hat{A}^{(m,i)}$ is the affinity matrix introduced in Eq.~\ref{eq:scaled_similarity}. To mitigate the bias toward longer warping paths (which can accumulate higher raw scores), we normalize $\gamma_{i,m}$ by the path length, resulting in the average alignment score:

\begin{equation}
\hat{\gamma}_{i,m} = \frac{1}{|W^{(m,i)}|} \gamma_{i,m},
\label{eq:normalized_alignment}
\end{equation}

where $|W^{(m,i)}|$ is the number of matched frame–sub-action pairs. Since the similarity values in $\hat{A}^{(i,m)}$ lie in $[0,1]$ (due to the sigmoid in Eq.~\ref{eq:scaled_similarity}), the normalized alignment score $\hat{\gamma}_{i,m}$ also lies in the range $[0,1]$.

Finally, we predict the class whose sub-action sequence best aligns—on average—with the observed video frames. This is done by selecting the class with the highest normalized alignment score:

\begin{equation}
\hat{y}_i = \arg\max_{c_m \in \mathcal{Y}} \; \hat{\gamma}_{i,m}.
\label{eq:final_prediction}
\end{equation}

\begin{table*}
  
  \centering
  \begin{tabular}{lrccc}
    \toprule \textbf{Method}                                       & \textbf{\#Param} & \textbf{Top-1 (\%)} $\uparrow$ & \textbf{Top-2 (\%)} $\uparrow$ & \textbf{Top-3 (\%)} $\uparrow$ \\
    \hline
    Random (10 Trials)                                             & -                & $20.81$             & $42.04$             & $62.50$             \\
   \textcolor{gray}{Human Evaluation (Oracle)}                                      & \textcolor{gray}{-}                & \textcolor{gray}{61.64}             & \textcolor{gray}{-}                   & \textcolor{gray}{-}                   \\
    \hline
    mPLUG-Owl-Video~\cite{ye2023mplug}                             & $7B$             & $19.49$             & -                   & -                   \\
    VideoChat2~\cite{li2023videochat}                              & $7B$             & $21.27$             & -                   & -                   \\
    VideoLLaMA~\cite{zhang2023videollama}                          & $8B$             & $22.71$             & -                   & -                   \\
    LLaVA-Next-Video~\cite{zhang2024video}                         & $7B$             & $22.90$             & -                   & -                   \\
    Qwen2--VL~\cite{wang2024qwen2vl}                                & $7B$             & $30.24$             & -                   & -                   \\
    X-CLIP-L/14-16F~\cite{ni2022xclip}                             & $0.6B$           & $16.26$             & $33.74$             & $49.89$             \\
    \hline
    SigLIP--so400m~\cite{Zhai2023SigmoidLF} (mean-pool) & $0.9B$           & $22.94$             & $42.20$             & $63.70$             \\
    + DTW Alignment (\textit{Ours})                                & $0.9B$           & $25.72$             & $45.99$             & $66.26$             \\
    \textbf{ActAlign (\textit{Ours})}                              & $0.9B$           & \boldmath{$30.40\pm0.11$}  & \boldmath{$53.01$}  & \boldmath{$70.27$}  \\
    \bottomrule
  \end{tabular}
  \vspace{-7pt}
    \caption{\textbf{Zero-shot classification results on ActionAtlas.} Our method achieves state-of-the-art Top-1, Top-2, and Top-3 accuracy, outperforming all baselines and billion-parameter video--language models without any video--text supervision. These results highlight the open-set recognition capability of image-text models and the effectiveness of structured sub-action alignment over flat representations such as mean-pooling. We use context-rich prompting strategy.
    }
  \label{tab:zeroshot_baselines}
\end{table*}

\section{Experiment}

\subsection{Experimental Setup}

\paragraph{Dataset}  
Our approach is dataset- and domain-agnostic, as it relies solely on language-driven sub-action generation from LLMs and cross-modal alignment without task-specific supervision. To rigorously evaluate its generality, we adopt ActionAtlas~\cite{salehi2024actionatlas}—to our knowledge, the most diverse and challenging benchmark for fine-grained action recognition across various domains. For each video \( V_i \), we retain its multiple-choice candidate set \( \{c_{i,1}, \dots, c_{i,M_i}\} \), and replace each class label with an LLM-generated sub-action sequence. (Dataset statistics is provided in the Appendix.)

\paragraph{Evaluation Metrics}
Following~\cite{Chen_2021_ICCV, 10.1007/978-3-031-78354-8_21}, we report Top-$k$ accuracy for $k\in\{1,2,3\}$:
\[
  \text{Top-}k = \frac{1}{N}\sum_{i=1}^{N} \mathbb{I}\Bigl(\text{rank}_{i}(\hat y
  _{i})\le k\Bigr),
\]
where $\mathbb{I}$ is the indicator function and $\text{rank}_{i}(\hat y_{i})$ is
the position of the ground‐truth label in the descending list of scores $\{\hat\gamma
_{i,1},\dots,\hat\gamma_{i,M_i}\}$. This accounts for typical cases where fine-grained actions are semantically similar and alignment scores are closely clustered, allowing improvement to be captured within a narrowed candidate set. 

\paragraph{Experimental Detail}
We use SigLIP--so400m (patch size 14, $d=384$, ~878M parameters). We apply a moving-average smoothing with a fixed window size of $w=30$ frames (1s @30 fps) to reduce transient noise and emphasize consistent motion patterns. All experiments run on a single NVIDIA RTX A5000 GPU (25 GB). Inference consists of embedding extraction followed by DTW alignment over $N$ class candidates, each with $M$ sub-actions, and $T$ video frames—yielding $\mathcal{O}(NMT)$ time. 
Following the zero‐shot protocol, no example videos or sub-action sequence from these classes are used for any training or tuning. We only use candidate action class names $c_m$ to prompt for sub-action generation.

\subsection{Experimental Result}

\paragraph{Zero-Shot Comparisons}  
We begin with a random choice baseline and a SigLIP zero-shot baseline using mean-pooled frame embeddings. Following ViFi-CLIP~\cite{rasheed2023vifi}, each video \( V_i \) is encoded as \( \bar{\mathbf{z}}_i = \frac{1}{T_i} \sum_{t=1}^{T_i} \mathbf{z}_i^t \), where mean-pooled embeddings $\bar{\mathbf{z}}$ is compared via cosine similarity to each name embedding \( \phi_t(c_j) \).

We further compare against open-source video–language models, including entries from the ActionAtlas leaderboard and fine-tuned CLIP variants. Despite using no video–text supervision, our method outperforms the SigLIP baseline by +7\% Top-1 and +11\% Top-2 accuracy, surpassing all baselines and billion-parameter models with $\sim$8$\times$ fewer parameters. These larger models are often optimized for interactive tasks through extensive instruction tuning, rather than for classifying fine-grained actions. In contrast, image–language models are trained on large-scale image–text pairs, making them particularly well-suited for open-set recognition when paired with temporal alignment mechanisms.

As shown in Table~\ref{tab:zeroshot_baselines}, these gains stem from our subaction-level alignment, which enables more discriminative and interpretable classification than global frame pooling method.

\paragraph{Ablation Studies}
We ablate each component of ActAlign on ActionAtlas (Table~\ref{tab:ablation}), starting from a mean-pooled SigLIP baseline. Adding DTW alignment introduces temporal structure and yields consistent gains. Context augmentation—injecting domain context (e.g., ``Sprint start'' $\rightarrow$ ``Sprint start in basketball'')—produces the largest boost by resolving semantic ambiguity. Signal smoothing offers a modest but complementary improvement by reducing frame-level noise and clarifying action boundaries. 

\begin{table}
  \centering

  \begin{tabular}{lccc}\toprule \textbf{Configuration} & \textbf{Top-1 (\%)} $\uparrow$ & \textbf{Top-2 (\%)}  $\uparrow$ & \textbf{Top-3 (\%)}  $\uparrow$ \\ \midrule SigLIP~\cite{Zhai2023SigmoidLF} (mean-pool) & $22.94$ & $42.20$ & $63.70$ \\ 
  + DTW Alignment & $25.72$ & $45.99$ & $66.26$ \\ 
  + Context Augmentation & $30.07$ & $52.67$ & $70.49$ \\ 
  + Signal Smoothing & $30.29$ & $53.01$ & $70.27$ 
 \\ \bottomrule\end{tabular} 
    \caption{\textbf{Ablation results of various design} under context-rich prompting. DTW alignment introduces temporal matching, context augmentation reduces sub-action ambiguity, and signal smoothing mitigates frame-level noise.}
  \label{tab:ablation}
\end{table}

\paragraph{Baseline Comparisons} To further evaluate the generalization power of SigLIP without any temporal information, we evaluate SigLIP performance by the similar context augmentation technique we used in our framework and with mean-pooling the bag of sub-actions and comparing the mean-pooled representations with video representation (See Table~\ref{tab:siglip_baselines}). In all of these baselines, video frame representations are mean-pooled. We further investigate classification performance under randomized and reversed sub-action orders at Table~\ref{tab:sub-action-baselines}

\begin{table}[t]
  \centering
  \begin{tabular}{lccc}
    \toprule
    \textbf{Configuration} & \textbf{Top-1 (\%)} $\uparrow$ & \textbf{Top-2 (\%)} $\uparrow$ & \textbf{Top-3 (\%)} $\uparrow$ \\
    \midrule
    SigLIP (mean-pool) & $22.94$ & $42.20$ & $63.70$ \\
    SigLIP (mean-pool) w/ Context Augmentation & $28.51$ & $50.89$ & $70.38$ \\
    SigLIP (mean-pool) w/ Bag-of-Words & $29.06$ & $51.34$ & $69.49$ \\
    \hline
    \textit{ActAlign (ours)} & $30.29$ & $53.01$ & $70.27$ \\
    \bottomrule
  \end{tabular}
  \caption{\textbf{Comparison with baselines under context-rich prompting.} Context augmentation enriches \textit{action names} (as opposed to sub-actions in Table~\ref{tab:ablation}) with domain names, while bag-of-words mean-pools sub-action descriptions into a unified representation.}
  \label{tab:siglip_baselines}
\end{table}

\paragraph{Sub-action Baselines} To evaluate the impact of sub-actions on alignment, we test DTW alignment performance under two perturbations: randomizing sub-actions and reversing their order on short-fixed sub-actions. Results are shown in Table~\ref{tab:sub-action-baselines}. 

\begin{table}
  \centering
  \begin{tabular}{llll}
    \toprule
    \textbf{Configuration} & \textbf{Top-1 (\%)} $\uparrow$ & \textbf{Top-2 (\%)} $\uparrow$ & \textbf{Top-3 (\%)} $\uparrow$ \\
    \midrule
    Reversed Order & $26.39$ & $47.66$ & $67.04$ \\
    Randomized Order (5 Trials) & $26.63\pm 0.14$ & $47.05\pm 0.29$ & $67.00\pm 0.25$ \\
    Normal & $27.06$ & $46.88$ & $66.82$ \\
    \bottomrule
  \end{tabular}
  \vspace{-8pt}
  \caption{\textbf{Comparison with sub-action baselines.} We evaluate classification performance by randomizing and reversing the order of \textit{short-fixed} sub-actions due to their larger sequence size.}
  \label{tab:sub-action-baselines}
\end{table}

We observe that the upper bound on performance is closely tied to the specificity and coherence of LLM-generated sub-action sequences.

\paragraph{Prompt Variations}

We evaluate two prompt strategies for generating sub-action scripts using GPT-4o, keeping all other components fixed:

\begin{itemize}
  \item \textbf{Short-Fixed:} Prompts GPT-4o to generate exactly 10 terse (2–3 word) sub-actions per class using a fixed structure.
  
  \item \textbf{Context-Rich:} Produces variable-length, context-rich sub-action scripts incorporating domain-specific cues (e.g., “Wrestler”, “Rider”).
\end{itemize}

\begin{table}
  \centering
    \begin{tabular}{llc}
      \toprule
      \textbf{Prompt} & \textbf{Description} & \textbf{Top-1 (\%)}  $\uparrow$\\
      \midrule
      Short‐fixed & 2-word, fixed 10 sub-actions & $27.06$ \\
      Context‐rich & context-rich, variable-length & $\textbf{30.29}$ \\
      \bottomrule
    \end{tabular}
   \vspace{-8pt}
  \caption{\textbf{Impact of prompt strategy.} Context-rich prompting improves zero-shot classification performance by producing more specific and informative sub-actions.}
  \label{tab:prompt_variations}
\end{table}

Table~\ref{tab:prompt_variations} shows the performance of each strategy. The context-rich prompt with domain-specific context achieve the highest accuracy. In contrast, short-fixed prompts—lacking sufficient semantic specificity—perform worst. These results highlight that reducing ambiguity in sub-action descriptions directly improves alignment quality and classification performance.

\begin{figure}
  \centering
  \begin{subfigure}[b]{0.37\linewidth}
    \centering
    \includegraphics[width=\linewidth]{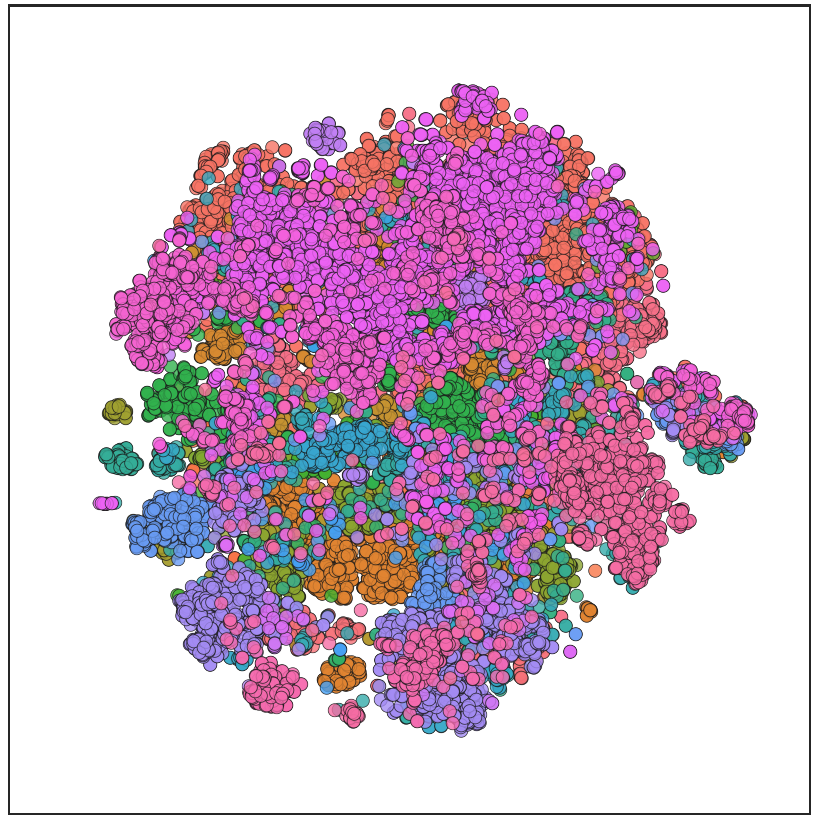}
    \caption*{\textbf{w/o Context Augmentation}}
  \end{subfigure}
  \hspace{0.01\linewidth} 
  \begin{subfigure}[b]{0.37\linewidth}
    \centering
    \includegraphics[width=\linewidth]{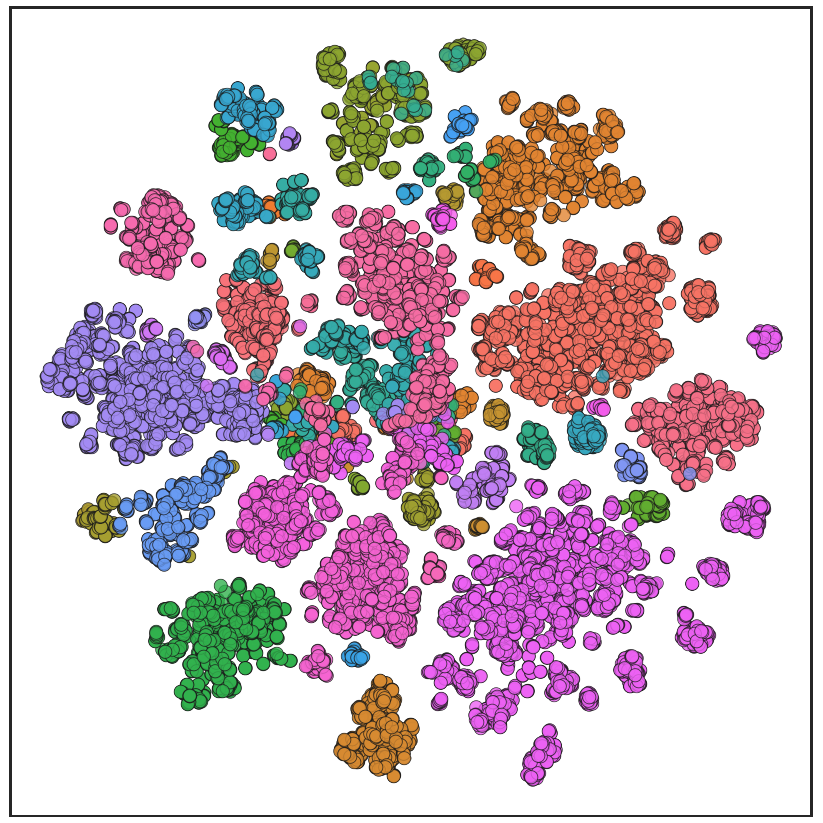}
    \caption*{\textbf{w/ Context Augmentation}}
  \end{subfigure}
  \vspace{-10pt}
  \caption{\textbf{t-SNE visualization of sub-action embeddings.} Each color corresponds to a sport domain. Augmenting sub-actions with context yields more discriminative clusters and improves textual grounding.}
  \label{fig:tsne_embeddings}
\end{figure}

\paragraph{Sub-action Embedding Structure}
Figure~\ref{fig:tsne_embeddings} shows 2D t-SNE projections of sub-action embeddings with and without context augmentation under context-rich prompts. Adding domain-specific cues (\texttt{<sub-action> in <sport name>}) results in tighter, more coherent clusters—indicating better semantic structure and separation.

\paragraph{Sub-action Sequence Examples}
Table~\ref{tab:example_scripts} shows LLM-generated sub-action sequences for two \textit{Figure skating}
tactics under our context-rich prompting. The scripts highlight ordered, salient steps enabling precise temporal matching.

\begin{table*}[t]
  \centering
  \renewcommand{\arraystretch}{1.1}
  \begin{tabular}{p{\textwidth}} 
    \toprule
    \textbf{Sub-action Script} \\ 
    \midrule

    \colorbox{mygray}{\textit{Biellmann Spin}} \\ 
    1. Begins upright spin on one foot with arms extended and free leg behind, 
    2. Gradually pulls free leg upward behind the back using both hands, 
    3. Raises the free leg above head level while arching the back dramatically, 
    4. Grasps the blade of the free skate with both hands overhead, 
    5. Extends spinning leg vertically while maintaining centered spin on skating foot, 
    6. Maintains high-speed rotation with body in extreme vertical split position \\ 
    \midrule

    \colorbox{mygray}{\textit{Flying Camel Spin}} \\ 
    1. Skater glides forward with arms extended and knees bent in preparation, 
    2. Performs a powerful jump off the toe pick while swinging free leg upward, 
    3. Rotates mid-air with body extended horizontally like a 'T' shape, 
    4. Lands on one foot directly into a camel spin position with torso parallel to ice, 
    5. Extends free leg backward and arms outward while spinning on the skating leg, 
    6. Maintains fast, centered rotation in the horizontal camel position \\ 

    \bottomrule
  \end{tabular} 
\vspace{-8pt}
    \caption{\textbf{LLM-generated sub-action scripts for figure skating tactics.} Shown for the \textit{Biellmann Spin} and \textit{Flying Camel Spin} examples in Figure~\ref{fig:heatmaps}, these sequences are generated using context-rich prompting and provide semantically detailed, temporally ordered steps for alignment in our zero-shot framework.}
  \label{tab:example_scripts}
\end{table*}

\begin{figure}[t]
  \centering

  \begin{subfigure}[b]{\linewidth}
    \centering
    \includegraphics[width=\linewidth]{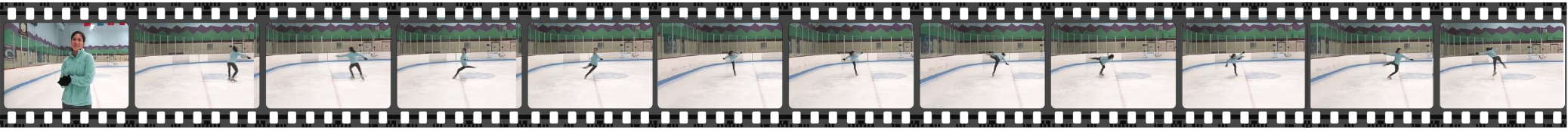}
    \caption{\textbf{Video:} Performing \emph{flying camel spin} in figure skating.}
    \label{fig:video_snapshot}
  \end{subfigure}

  \begin{subfigure}[b]{0.39\linewidth}
    \centering
    \includegraphics[width=\linewidth]{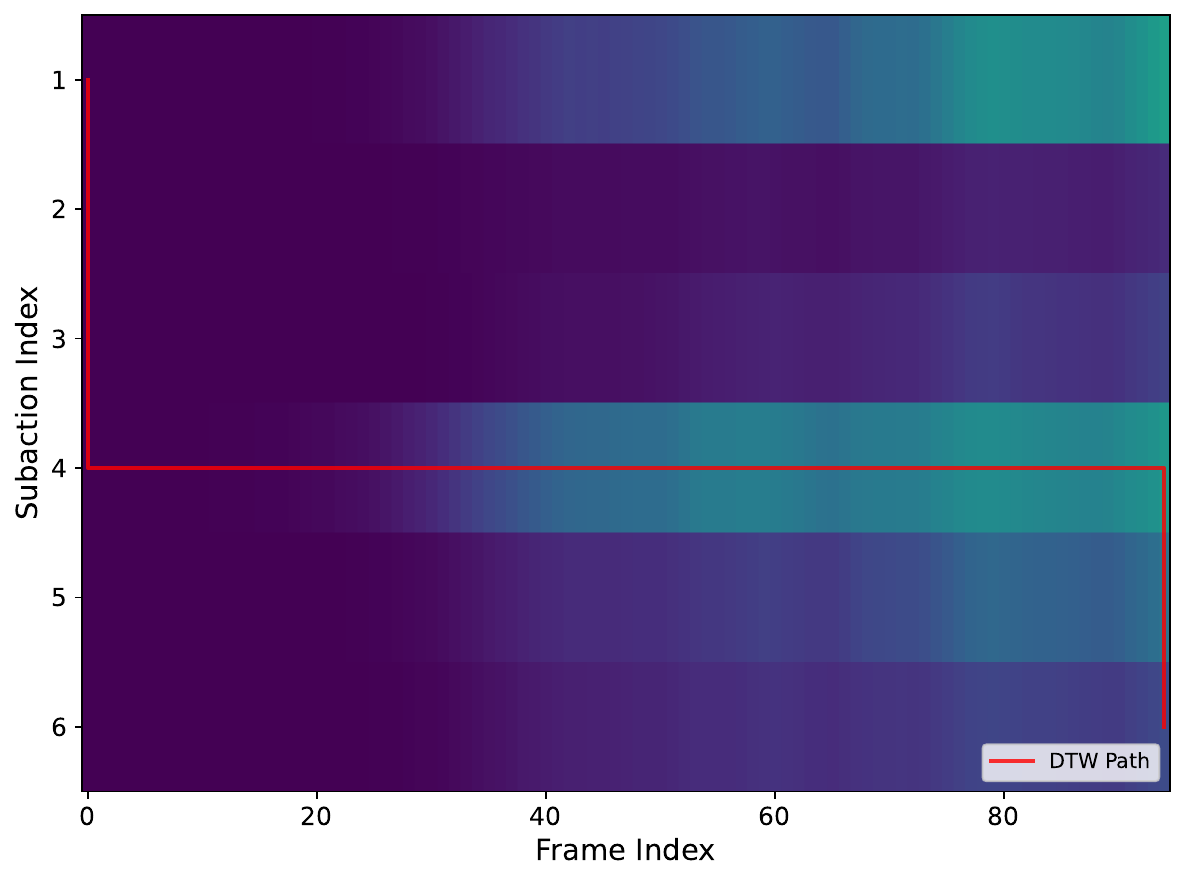}
    \caption{Biellmann spin \textcolor{red}{$\times$}}
    \label{fig:heatmap_incorrect}
  \end{subfigure}
  \hspace{0.04\linewidth}
  \begin{subfigure}[b]{0.39\linewidth}
    \centering
    \includegraphics[width=\linewidth]{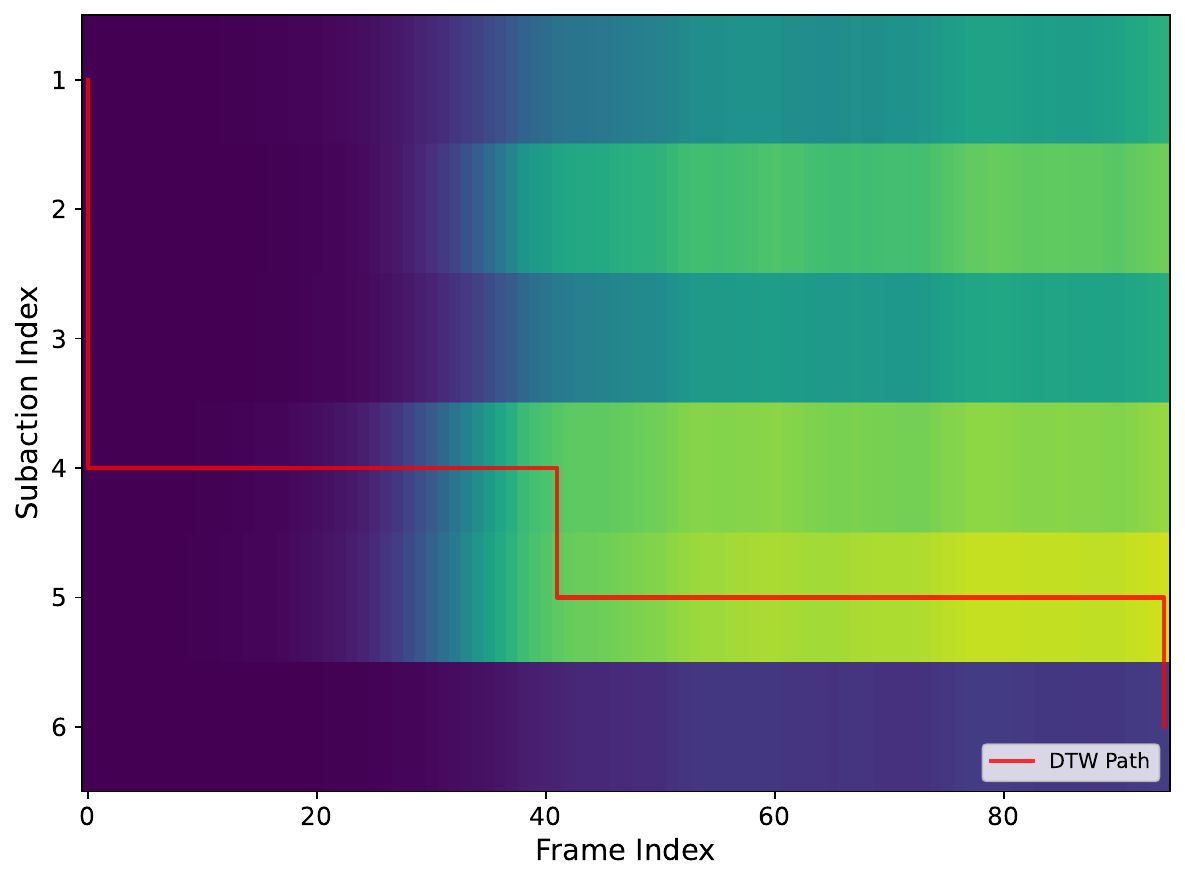}
    \caption{Flying camel spin \textcolor{green}{$\checkmark$}}
    \label{fig:heatmap_correct}
  \end{subfigure}

\vspace{-10pt}
  \caption{
  \textbf{DTW alignment paths} for an incorrect prediction (left) versus a correct classification (right). The correct class exhibits clearer segmentation and higher alignment quality. The sub-action scripts are provided in Table~\ref{tab:example_scripts}.
  }
  \label{fig:heatmaps}
\end{figure}

\paragraph{Alignment Heatmaps and Paths}
Figure~\ref{fig:heatmaps} visualizes the cosine similarity matrix and DTW path for a correctly classified action (right) and an incorrect candidate (left). The correct sequence yields high-similarity regions with a monotonic path. In contrast, the incorrect script shows sparse similarity regions, with the DTW path forced to follow the single most similar alignment trace.



\begin{figure}[t]
  \centering

  \begin{subfigure}[b]{0.39\linewidth}
    \centering
    \includegraphics[width=\linewidth]{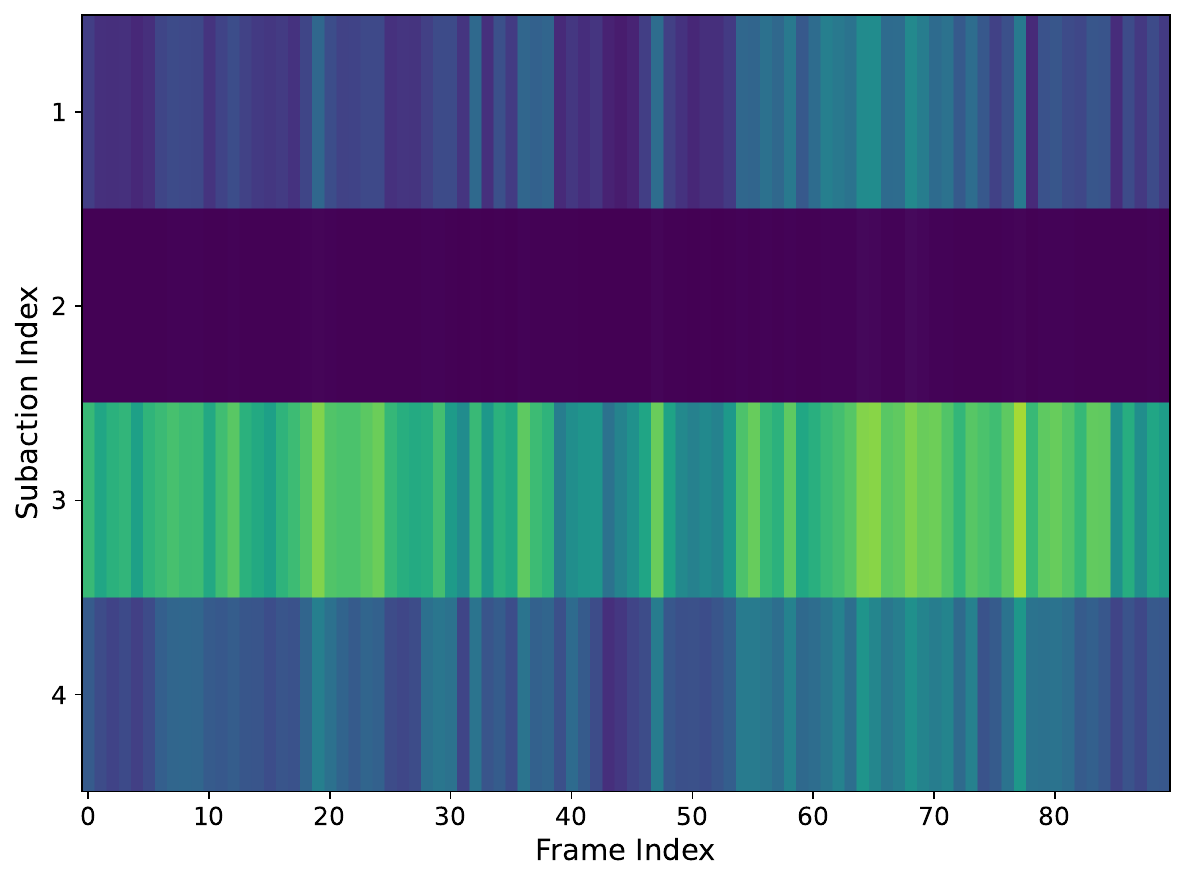}
    \caption{w/o smoothing}
    \label{fig:sim_wo_smoothing}
  \end{subfigure}
  \hspace{0.04\linewidth}
  \begin{subfigure}[b]{0.39\linewidth}
    \centering
    \includegraphics[width=\linewidth]{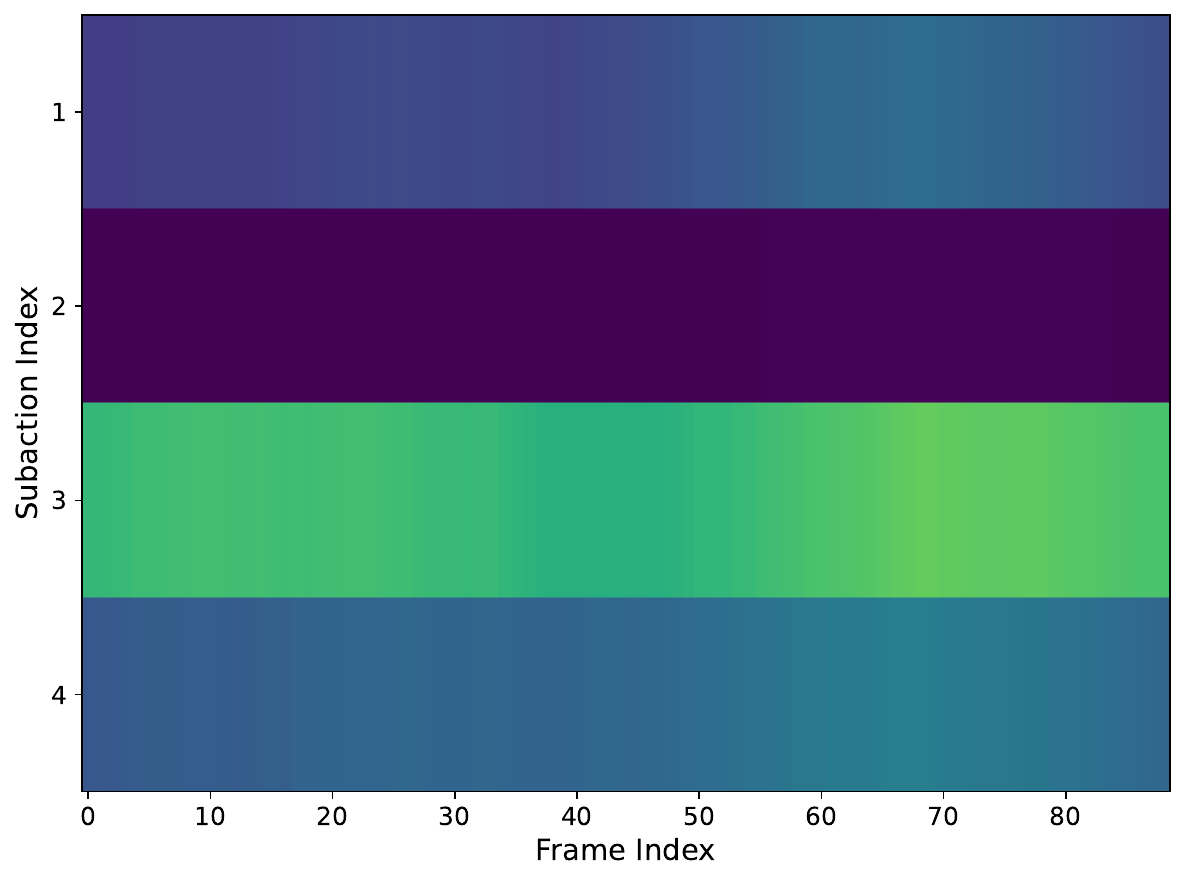}
    \caption{w/ smoothing 
    }
    \label{fig:sim_w_smoothing}
  \end{subfigure}
\vspace{-10pt}
    \caption{\textbf{Signal smoothing} reduces high-frequency noise and enhances transition between sub-actions. Similarity matrices before and after applying a moving-average filter ($w=30$). }
  \label{fig:sim_matrix_smoothing}
\end{figure}

\paragraph{Signal Smoothing.}
Figure~\ref{fig:sim_matrix_smoothing} compares similarity matrices before and after applying a moving-average filter. Without smoothing, rapid scene changes introduce noise and scattered peaks. Smoothing yields cleaner similarity surfaces with clearer action boundaries.

\begin{figure}
  \centering

  \begin{subfigure}[b]{\linewidth}
    \centering
    \includegraphics[width=1.\linewidth]{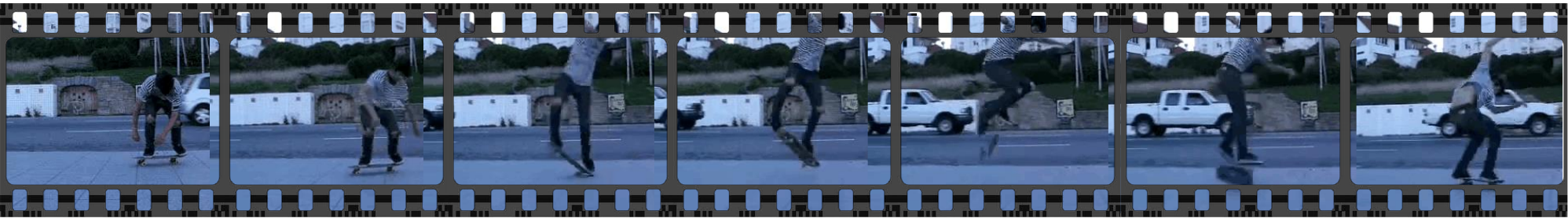}
    \caption{\textbf{Video:} 
    \emph{Varial kickflip underflip body varial} in Skateboarding.}
    \label{fig:snapshot_clinch}
  \end{subfigure}

  \vspace{0.5em}

  \begin{subfigure}[b]{0.39\linewidth}
    \centering
    \includegraphics[width=\linewidth]{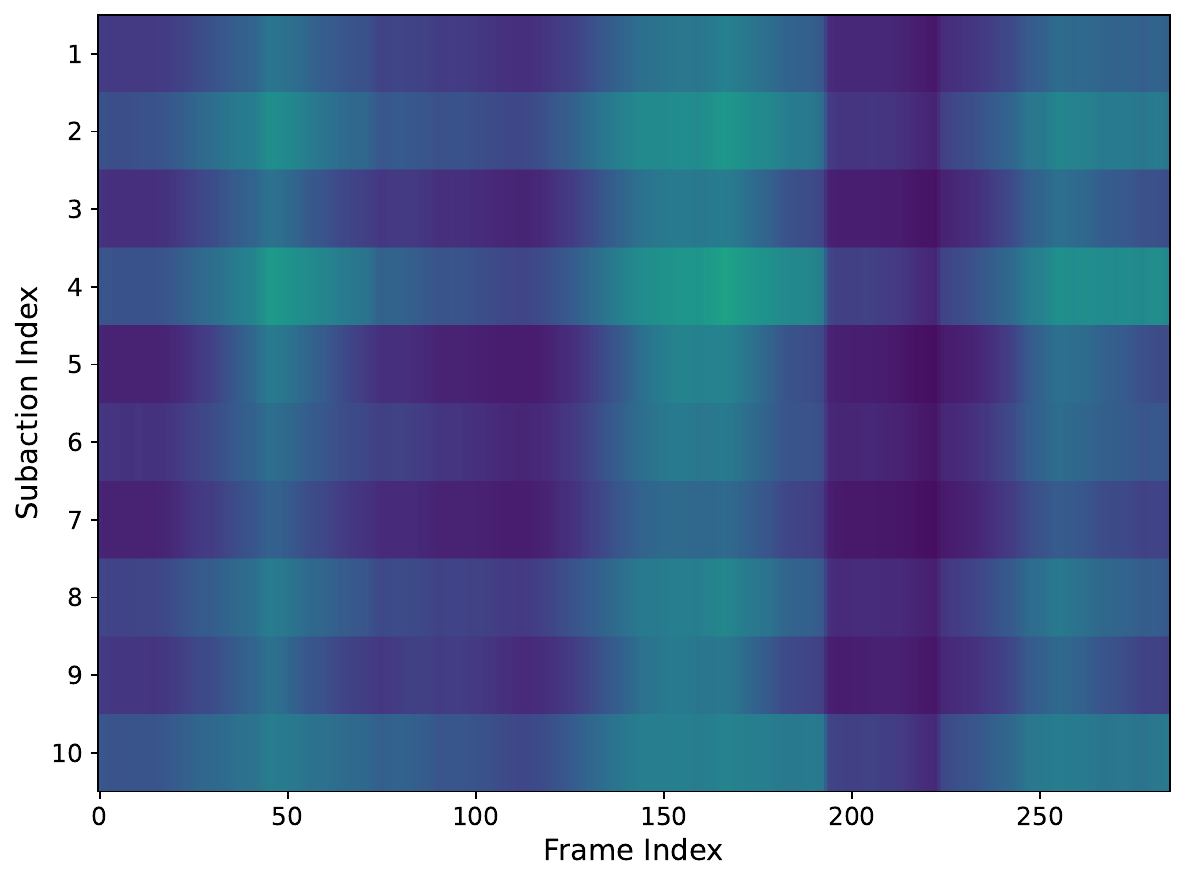}
    \caption{Short‐fixed (\textcolor{red}{$\times$}, $\hat{\gamma}=0.35$).}
    \label{fig:heatmap_sf}
  \end{subfigure}
  \hspace{0.04\linewidth}
  \begin{subfigure}[b]{0.39\linewidth}
    \centering
    \includegraphics[width=\linewidth]{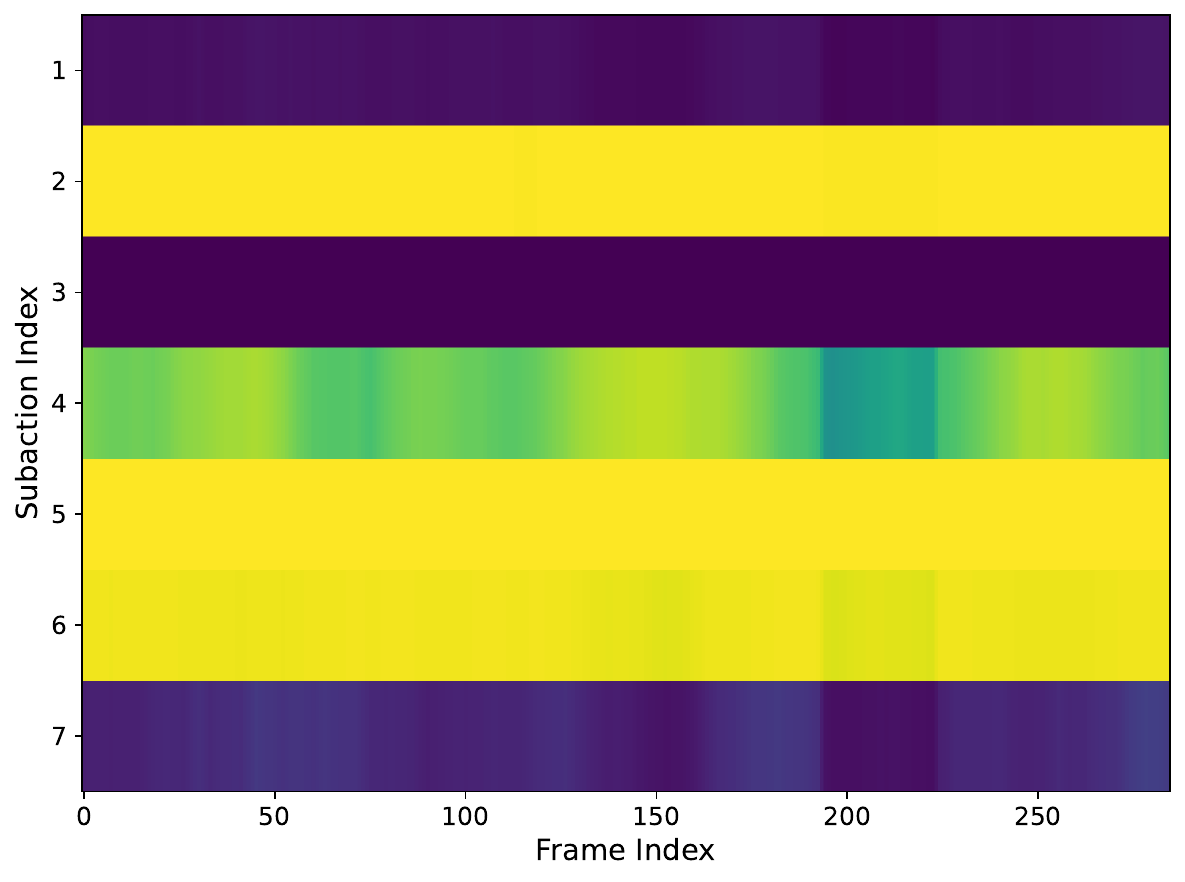}
    \caption{Context‐rich (\textcolor{green}{$\checkmark$}, $\hat{\gamma}=0.81$).}
    \label{fig:heatmap_cr}
  \end{subfigure}
    \vspace{-8pt}\caption{\textbf{Failure case} due to vague sub-actions for the action \emph{Varial kickflip underflip body varial}. The normalized alignment score $\hat{\gamma}$ reflects overall sequence similarity. (\subref{fig:heatmap_sf}) The Short-fixed prompt yields a noisy similarity map and incorrect prediction. (\subref{fig:heatmap_cr}) The Context-rich prompt produces clearer alignment and correct classification. 
    See Appendix for the detailed scripts.
    }
    \label{fig:heatmap_comparison}
\end{figure}

\paragraph{Failure Case Analysis}  
We observe two primary failure modes in our framework:

\begin{itemize}
  \item \textbf{Ambiguous sub-actions:} Vague LLM-generated steps (e.g., “move to position”) result in sparse similarity matrices, weakening DTW’s ability to discriminate between candidates. As shown in Figure~\ref{fig:heatmap_comparison}, context-rich prompts produce clearer alignment regions than short-fixed ones.

  \item \textbf{Global alignment bias:} Vanilla DTW enforces full-sequence alignment, which is suboptimal when actions start mid-clip or exhibit temporal shifts. Without a local alignment mechanism, early or trailing sub-actions can introduce noise. Figure~\ref{fig:heatmaps} shows a temporal shift case that does not affect prediction here but may cause errors in more competitive settings.
\end{itemize}

These limitations underscore the importance of precise sub-action design and motivate future improvements in alignment robustness and prototype refinement.

\section{Conclusion}
We show that contrastive image–language models establish a surprisingly strong baseline for zero-shot fine-grained video classification, even when used with simple mean-pooling. To fully leverage this capability in extremely fine-grained settings, we propose \textbf{ActAlign}, a novel \emph{zero-shot, domain-general, and model-agnostic} framework that revisits the classic Dynamic Time Warping (DTW) algorithm to cast video classification as a sequence-alignment problem. By aligning video frames with LLM-generated sub-action scripts, ActAlign introduces temporal structure into contrastive models without requiring any video–text training or fine-tuning. Evaluated on the highly challenging and diverse ActionAtlas~\cite{salehi2024actionatlas} benchmark, our method achieves state-of-the-art performance, outperforming both CLIP-style baselines and billion-parameter video–language models. 


\bibliography{mybibfile}
\bibliographystyle{tmlr}

\clearpage
\appendix
\newcommand{\maketitlesupplementary}{%
  \maketitle
  \begin{center}
    {\large \bf ActAlign: Zero-Shot Fine-Grained Video Classification
via Language-Guided Sequence Alignment}
  \end{center}
  \vspace{1em}
}

\section{Smoothing Analysis}
We assess the effect of signal smoothing on model performance. Specifically, we vary the moving-average window size \( w \in \{10, 20, 30, 50\} \), applied over the frame-wise SigLIP embeddings before alignment.

\begin{figure}[h]
  \centering
  \includegraphics[width=0.60\linewidth]{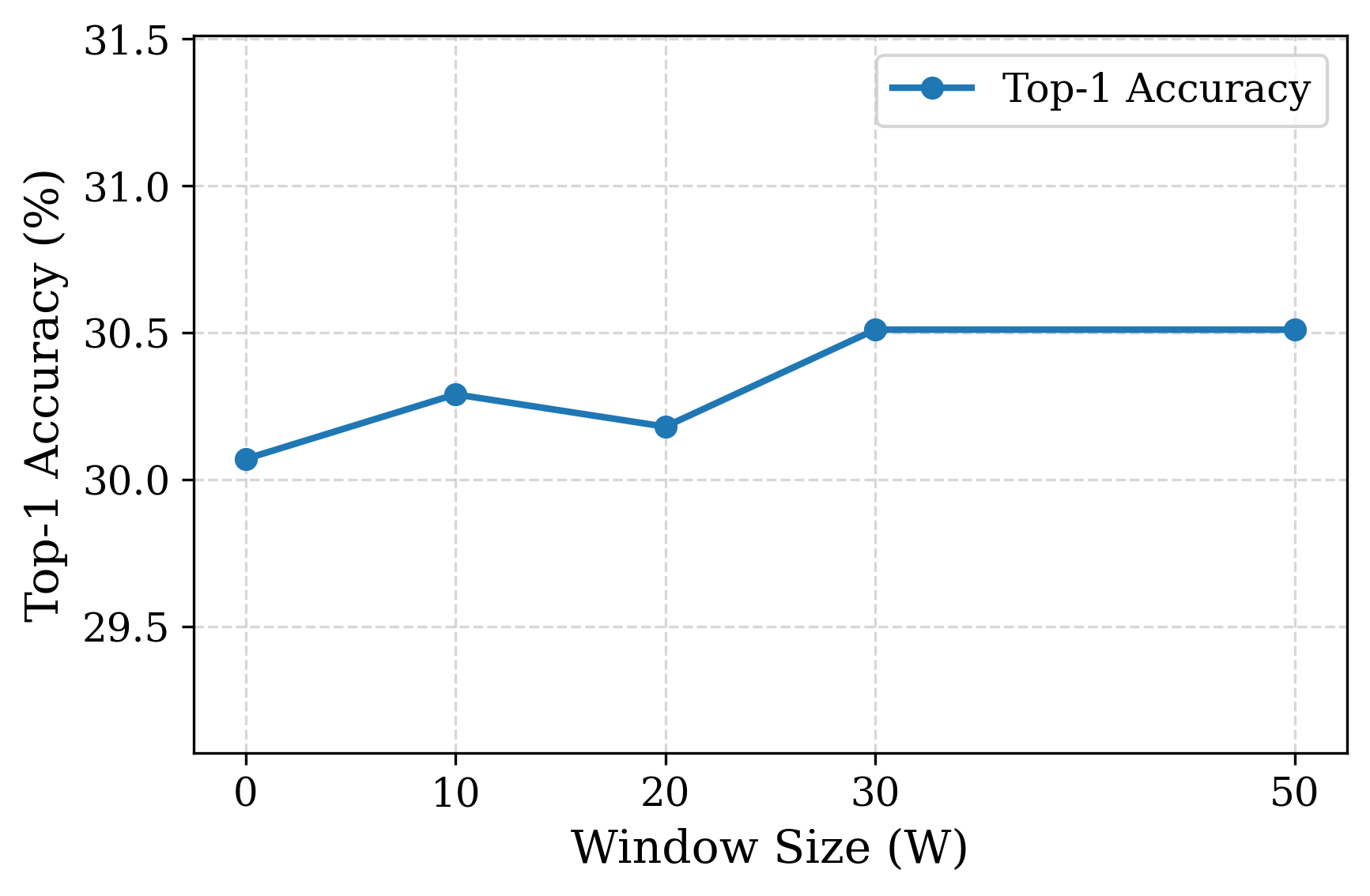}
  \caption{Top-1 accuracy for different smoothing window sizes $w$. Accuracy is stable across settings, with $w=30$ selected as default.}
  \label{fig:smoothing_sweep}
\end{figure}

\paragraph{Effect on Accuracy.}
As shown in Figure~\ref{fig:smoothing_sweep}, the Top-1 classification accuracy remains almost stable across a wide range of smoothing widths. This suggests that our method is not overly sensitive to the smoothing parameter.

\paragraph{Choice of Default.}
We fix \( w = 30 \), which corresponds to a 1-second temporal window under 30 FPS videos. This setting balances local context preservation with jitter reduction and yields smooth DTW alignments that more cleanly segment sub-actions.

\section{Implementation Details and Inference Pipeline}

\subsection{System Configuration}
We conduct all experiments on a server running Ubuntu 20.04 with a single NVIDIA RTX A5000 GPU (24GB VRAM) and 256GB RAM. The pipeline is implemented in PyTorch 2.7 with CUDA 12.8. For subaction generation, we use the GPT-4o API.

\subsection{Model Settings}
We use the publicly available SigLIP model from Google (\texttt{siglip-so400m-patch14-384}), which supports variable input resolutions. Images are embedded using a ViT-based vision encoder with a patch size of \(14 \times 14\) and an output embedding dimension of \(d = 384\). Sub-actions are encoded using the corresponding frozen SigLIP text encoder. All embeddings are \(\ell_2\)-normalized before computing cosine similarities.

\subsection{Runtime Performance}
End-to-end inference over all 898 samples—including affinity matrix computation, signal smoothing, temporal alignment, and classification—completes in 41.95 seconds, with an average processing time of 0.04 seconds per video. Dynamic time warping (DTW) alignment accounts for over 90\% of the total runtime.

\subsection{Scalability Analysis} 
From an algorithmic standpoint, the scalability is primarily determined by the dynamic time warping (DTW) alignment between the visual frame sequence and sub-action scripts. The overall complexity scales as $\mathcal{O}(NMT)$, where $N$ is the number of candidate classes, $M$ is the average number of sub-actions per class, and $T$ is the number of video frames. In practice, $M$ remains small (on the order of tens) and independent of video length. Because the method operates entirely in a frozen embedding space and requires no gradient updates, it avoids the substantial memory and computational overhead associated with training large video-language models. As a result, \textsc{ActAlign} scales gracefully to longer videos and larger label vocabularies. Moreover, since alignment computations for different candidate classes and frame segments are independent, the framework is naturally amenable to parallelization across GPUs or distributed compute resources, offering a straightforward path for further acceleration in large-scale or real-time settings.

\subsection{Inference Pipeline Overview}
Our pipeline proceeds in six modular stages: (1) videos from ActionAtlas are downloaded and clipped using annotated timestamps, with frames stored in a structured NumPy archive; (2) each frame is encoded using SigLIP’s vision encoder to obtain per-frame embeddings; (3) GPT-4o generates context-rich sub-action sequences for each candidate label using fixed prompt templates; (4) sub-actions are encoded via SigLIP’s text encoder; (5) cosine similarity matrices are computed between visual and semantic sequences, and DTW is applied to compute alignment scores for each candidate; and (6) as a baseline, we compare mean-pooled frame embeddings with text embeddings of class labels using cosine similarity, bypassing alignment. Each step is modular and implemented for fast inference, allowing to easily switch between models and datasets. 

\section{More Related Works}
\textbf{Image–Language Transfer for Video Tasks}. Image–language models pretrained on large-scale image–text pairs exhibit strong open-set recognition capabilities. This motivates their adaptation for video tasks. Common strategies include adding temporal modules~\cite{wang2022actionclip, ni2022xclip, rasheed2023vifi, liu2024omniclipadaptingclipvideo} or introducing learnable prompts~\cite{ju2022prompt, lin2022evl}. However, as noted by~\cite{rasheed2023vifi}, fine-tuning often compromises the model's open-set generalization. To preserve zero-shot capabilities, recent work explores training-free adaptations using temporal prompting methods~\cite{ahmad2024ezclipefficientzeroshotvideo, Phan_2024_WACV, Yu_Cao_Zhang_Lv_Min_Zhang_2025}. For example, TEAR~\cite{10.1007/978-3-031-78354-8_21} compares mean-pooled frame embeddings with LLM-generated action descriptions for each class. While effective, these approaches either degrade generalization through fine-tuning or ignore temporal structure via mean-pooling - limiting performance on fine-grained and zero-shot video classification.

\section{Prompt Templates and Generation Settings}

We use GPT-4o to automatically convert each high-level class name into a sequence of sub-actions. Two prompt styles were explored:

\subsection{Short-Fixed Prompt.}  
This setting produces terse outputs (e.g., “tie arms,” “deliver knee”) that often lack domain-specific context. The descriptions are directly adopted from the orginal ActionAtlas~\cite{salehi2024actionatlas} dataset.  

\subsubsection{Template}
\begin{mdframed}[backgroundcolor=lightgray, linecolor=lightgray, innertopmargin=4pt, innerbottommargin=4pt, innerleftmargin=6pt, innerrightmargin=6pt]\small\textit{%
Generate a JSON object from the following context.\\[1ex]
For each action option given below, create a key in the JSON object with that action name.\\
The value for each key must be a list of exactly 10 subactions.\\
Each subaction must be a minimalist, two‐word description.\\
The descriptions should capture the essential mechanics of the action.\\[1ex]
Use the following action options and their descriptions for context:\\[.5ex]
- \texttt{<action 1>}: \texttt{<action 1 description>}\\
- \texttt{<action 2>}: \texttt{<action 2 description>}\\
- \texttt{<action 3>}: \texttt{<action 3 description>}\\
\ldots\\[1ex]
Output the JSON object accordingly.%
}
\end{mdframed}

\subsection{Context-Rich Prompt.} 
This setting encourages richer descriptions with more related keywords grounded in the activity domain. Context-rich prompting strategy significantly improves textual grounding during DTW alignment.

\subsubsection{Template}
\begin{mdframed}[backgroundcolor=lightgray, linecolor=lightgray, innertopmargin=4pt, innerbottommargin=4pt, innerleftmargin=6pt, innerrightmargin=6pt]
\small\textit{%
You will output a JSON object. Each key is an action name from the list below, and its value is a list of concise, visually distinctive subaction descriptions performed sequentially.\\[1ex]
Do NOT fix the list length—choose however many substeps best break down that trick into discriminative subactions.\\[1ex]
Each subaction should be:\\
\hspace*{1em}– Self-contained and visually descriptive with rich \texttt{<domain>} keywords and context.\\
\hspace*{1em}– Explicitly reference the scene, objects, environment, and motion in the given sport context.\\
\hspace*{1em}– Discriminative and concise.\\[1ex]
Here are the actions to decompose:\\
– \texttt{<action 1>} in \texttt{<domain>}\\
– \texttt{<action 2>} in \texttt{<domain>}\\
– \texttt{<action 3>} in \texttt{<domain>}\\
\ldots\\[1ex]
Output ONLY the JSON object, without any additional explanation.%
}
\end{mdframed}

\subsection{Examples}
Table~\ref{tab:script_comparison} compares the two prompting strategies for a skateboarding trick example. The corresponding affinity matrices are shown in Figure~\ref{fig:heatmap_comparison}.

\paragraph{Note.} Our experiments demonstrate that context-rich prompts significantly outperform short-fixed ones, as the added context helps disambiguate tactics across sports and yields smoother DTW alignment with video frames.

\begin{table}[t]
  \centering
  {\small 
  \begin{tabular}{l|p{0.5\linewidth}} \hline 
  \textbf{Prompt Variant} & \textbf{Generated Subaction Script} \\ \hline 
  
  Short‐fixed & 
  1. flip board \newline 
  2. spin board \newline 
  3. rotate body \newline 
  4. catch board \newline 
  5. kick upward \newline 
  6. flip reverse \newline 
  7. control motion \newline 
  8. hover steady \newline 
  9. anticipate landing \newline 
  10. land cleanly \\ \hline 
  
  Context‐rich & 
  1. Rider rolls forward on flat ground, knees compressed and eyes focused ahead. \newline
  2. Back foot scoops the tail into a backside shuvit while front foot flicks a kickflip off the nose-side edge. \newline
  3. As the board begins its varial kickflip rotation, the front foot quickly retracts and then kicks downward under the board’s underside. \newline
  4. Board completes a varial kickflip and then reverses flip direction mid-air due to the underflip, creating a complex double-flip motion. \newline
  5. Simultaneously, rider initiates a 180-degree body varial, rotating in the same direction as the board’s shuvit. \newline
  6. Rider tracks the board’s griptape, completes the body spin, and catches the board with both feet aligned over the bolts. \newline
  7. Rider lands smoothly, now facing the opposite direction, and rolls away with balance. \\ \hline
  \end{tabular}
  }
  \caption{Sub-action sequence comparison for the \textit{varial kickflip underflip body varial trick} in skateboarding using different prompt variants. Figure~\ref{fig:heatmap_comparison} represents the heatmap comparisons.}
  \label{tab:script_comparison}
\end{table}

\section{Dataset Statistics}
\paragraph{Dataset Overview.}  
Table~\ref{tab:dataset_stats} summarizes key statistics of the ActionAtlas~\cite{salehi2024actionatlas} dataset, which we extended to enable alignmnet-based classification. The dataset contains 898 video clips spanning 558 unique tactics across 56 sports domains. Each video is paired with 4 to 6 fine-grained candidate actions (5.25 in average per video). Videos vary significantly in length, with a mean of 173 frames (30 FPS), ranging from extremely short clips (3 frames) to multi-second sequences (up to 900 frames). Notably, ActionAtlas is the most diverse and challenging fine-grained action recognition benchmark to the best of our knowledge. It poses a major challenge in fine-grained recognition: human top-1 accuracy is just 61.64\%.

\begin{table}[t]
  \centering
  \renewcommand{\arraystretch}{0.98}
  \begin{tabular}{p{0.25\textwidth} p{0.5\textwidth} p{0.2\textwidth}}
    \toprule
    \textbf{Category} & \textbf{Statistic} & \textbf{Value} \\
    \midrule
    \multirow{6}{=}{\textit{Dataset scale}} 
      & Videos                                   & 898 \\
      & Unique fine-grained actions              & 558 \\
      & Sport domains                            & 56 \\
      & Avg. candidate actions per video         & 5.26 ± 1.27 \\
      & Avg. candidate actions per domain        & 33.6 \\
      & Min–Max candidate actions per domain     & 4–201 \\
    \addlinespace
    \multirow{2}{=}{\textit{Temporal properties}} 
      & Avg. video length (frames)               & 173.2 ± 121.3 \\
      & Min–Max video frames                     & 3–900 \\
    \addlinespace
    \multirow{1}{=}{\textit{Recognition accuracy}} 
      & Human top-1 accuracy (\%)                & 61.64 \\
    \bottomrule
  \end{tabular}
  \caption{\textbf{Summary of ActionAtlas dataset statistics.}}
  \label{tab:dataset_stats}
\end{table}

\begin{figure}[ht]
    \centering
    \includegraphics[width=0.7\linewidth]{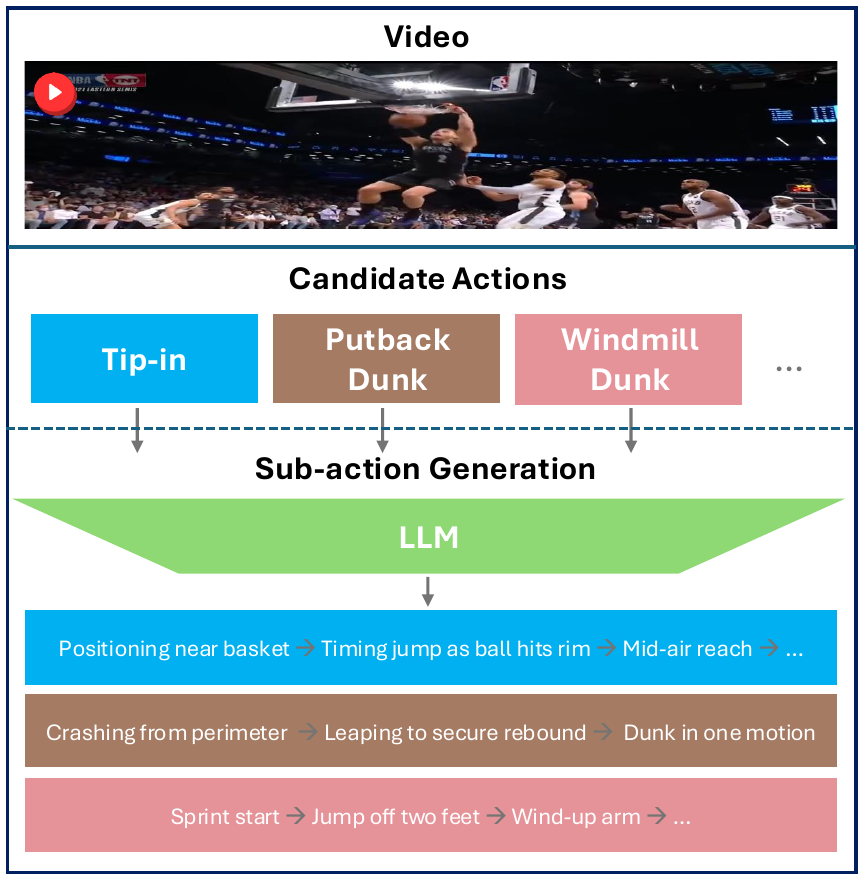}
    \caption{%
        Our pipeline uses an LLM to generate \textbf{structured sub-action sequences} for each fine-grained candidate action in ActionAtlas~\cite{salehi2024actionatlas} data sample. This structured representation enables sequence alignment with video frames for zero-shot recognition.
    }
    \label{fig:subactionatlas}
\end{figure}

\paragraph{ActionAtlas Extension.}
To enable subaction-level alignment, we augment the original ActionAtlas dataset with sub-action scripts generated by a large language model for each candidate action (see Figure~\ref{fig:subactionatlas}). This transforms the task from multiple-choice classification into alignment-based classification without modifying the original videos or labels. The extended dataset supports zero-shot video classification via sub-action sequence matching.

\paragraph{Sub-action Script Statistics.}
We generate sub-action sequences using different prompting strategies. As summarized in Table~\ref{tab:prompt_variants}, context-rich prompts yield fewer but more descriptive sub-actions, while short-fixed prompts produce uniform-length, terse sequences. These linguistic differences influence the textural grounding, and consequently, sequence alignment.

\begin{table*}[t]
  \centering
  \renewcommand{\arraystretch}{1.1}
  \begin{tabular}{p{0.25\textwidth} p{0.25\textwidth} p{0.25\textwidth} p{0.2\textwidth}}
    \toprule
    \textbf{Prompt Type} & \textbf{Avg. Subactions} & \textbf{Avg. Subactions per Domain} & \textbf{Avg. Words per Subaction} \\
    \midrule
    Short-fixed    & 10.00 ± 0.00   & 10.00 ± 0.00   & 2.00 ± 0.03 \\
    Context-rich   & 4.94 ± 0.86    & 5.01 ± 0.62    & 13.68 ± 2.78 \\
    \bottomrule
  \end{tabular}
  \caption{\textbf{Linguistic complexity of subaction scripts generated by different prompting strategies.} Context-rich prompts yield fewer but more descriptive subactions with significantly higher word counts compared to the Short-fixed strategy.}
  \label{tab:prompt_variants}
\end{table*}

\section{Performance within Domains}
We evaluate our approach on ActionAtlas, which covers over 50 sports domains while including extremely fine-grained actions, to assess its generality. Specifically, we focus on the four largest domains in the dataset and compare performance against the SigLIP baseline (Table~\ref{tab:topk_comparison_domains}). Our method consistently outperforms the baseline across all domains.

\begin{table*}[t]
\centering
\renewcommand{\arraystretch}{1.2}
\setlength{\tabcolsep}{5.5pt}
\begin{tabular}{l c c c c c c c c c c}
\hline
\multirow{2}{*}{\textbf{Domain}} & \multirow{2}{*}{\textbf{Samples}} 
& \multicolumn{3}{c}{\textbf{ActAlign (Ours)}} 
& \multicolumn{3}{c}{\textbf{SigLIP Baseline}} 
& \multicolumn{3}{c}{\textbf{$\Delta$ (pp)}} \\
\cline{3-5} \cline{6-8} \cline{9-11}
 & & Top-1 & Top-2 & Top-3 & Top-1 & Top-2 & Top-3 & Top-1 & Top-2 & Top-3 \\
\hline
Soccer       & 168 & \textbf{27.38} & \textbf{52.98} & \textbf{70.83} & 23.81 & 45.83 & 63.10 & +3.57 & +7.15 & +7.73 \\
Basketball   & 106 & \textbf{36.79} & \textbf{54.72} & \textbf{70.75} & 17.92 & 41.51 & 58.49 & +18.87 & +13.21 & +12.26 \\
Cheerleading &  61 & \textbf{22.95} & \textbf{55.74} & \textbf{67.21} & 22.95 & 37.70 & 60.66 & +0.00 & +18.04 & +6.55 \\
Wrestling    &  60 & \textbf{26.67} & \textbf{48.33} & 61.67          & 16.67 & 36.67 & \textbf{66.67} & +10.00 & +11.66 & -5.00 \\
\hline
\textbf{Mean} & --  & \textbf{28.45} & \textbf{52.94} & \textbf{67.62} & 20.34 & 40.43 & 62.23 & \textbf{+8.11} & \textbf{+12.52} & \textbf{+5.39} \\
\hline
\end{tabular}
\caption{\textbf{Top-$k$ accuracy (\%) comparison between ActAlign and SigLIP baseline across four largest ActionAtlas domains.} ActAlign demonstrates domain-agnostic improvements across all $k$ levels. $\Delta$ denotes the absolute percentage-point gain of ActAlign over the baseline. Bold indicates the higher result for each domain and metric.}
\label{tab:topk_comparison_domains}
\end{table*}

\section{Additional Qualitative Example}
\begin{figure}
  \centering

  \begin{subfigure}[b]{\linewidth}
    \centering
    \includegraphics[width=\linewidth]{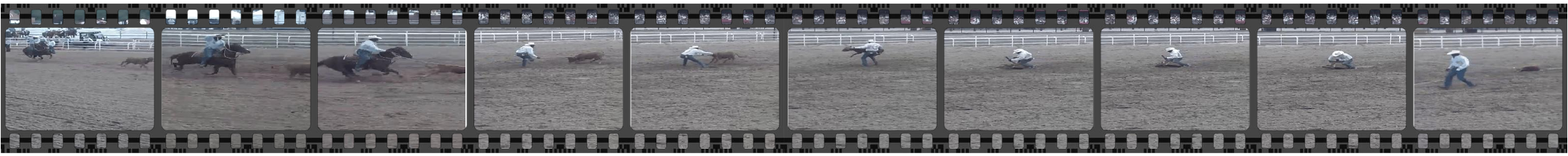}
    \caption{\textbf{Video:} Performing \emph{tie-down roping} in Rodeo.}
    \label{fig:video_snapshot2}
  \end{subfigure}

  \begin{subfigure}[b]{0.40\linewidth}
    \centering
    \includegraphics[width=\linewidth]{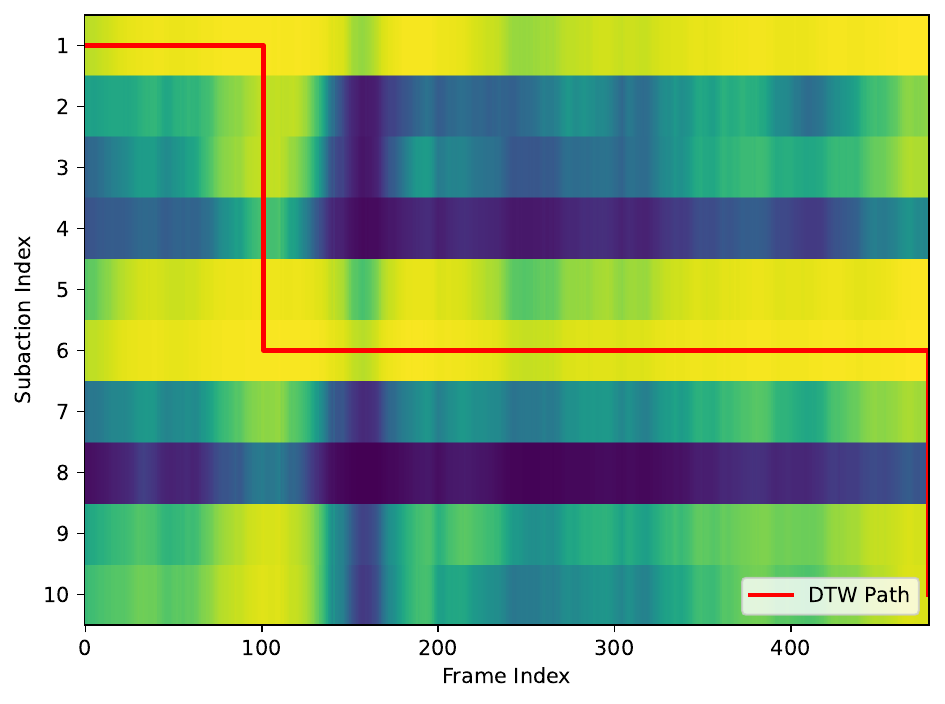}
    \caption{Short-fixed sub-actions, $\hat{\gamma}=0.96$ \textcolor{green}{$\checkmark$}}
  \end{subfigure}
  \hspace{0.04\linewidth}
  \begin{subfigure}[b]{0.40\linewidth}
    \centering
    \includegraphics[width=\linewidth]{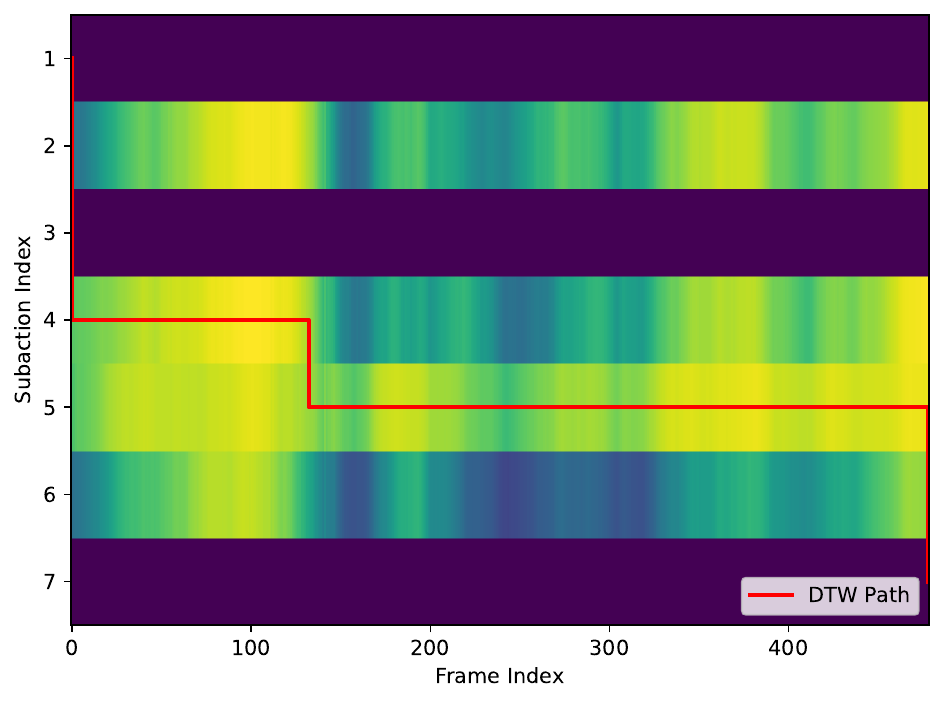}
    \caption{Context-rich sub-actions, $\hat{\gamma}=0.84$ \textcolor{green}{$\checkmark$}}
  \end{subfigure}

  \caption{
  \textbf{DTW alignment paths} comparison for a correctly predicted sample between short-fixed and context-rich sub-actions. The sub-action scripts are provided in Table~\ref{tab:SF-and-CR-comparison}.}
  \label{fig:SF-and-CR-comparison}
\end{figure}

To illustrate how different phrasings of sub-actions can influence alignment behavior, we present an example of \textit{tie-down roping} trick in Rodeo. Figure~\ref{fig:SF-and-CR-comparison} shows the corresponding DTW alignment paths, and Table~\ref{tab:SF-and-CR-comparison} lists the sub-action scripts used in each case.

\begin{table*}[t]
  \centering
  \renewcommand{\arraystretch}{1.15}
  \setlength{\tabcolsep}{6pt}
  \begin{tabular}{p{0.12\textwidth} p{0.36\textwidth} p{0.44\textwidth}}
    \toprule
    \textbf{Step} & \textbf{Short-Fixed Script} & \textbf{Context-Rich Script} \\
    \midrule

    1 & Pursue calf & Cowboy and horse launch from the roping box as the calf bolts across the arena. \\
    2 & Swing lasso & Roper swings the rope fluidly and hurls it in a clean loop that snags the calf's neck mid-gallop. \\
    3 & Catch neck & Rope tightens as horse halts and backs to maintain tension, stopping the calf abruptly. \\
    4 & Dismount quickly & Cowboy dismounts in one jump while horse holds the rope taut, sprinting to the flailing calf. \\
    5 & Reach calf & Flips the calf onto its side with a quick lift-and-roll maneuver in the loose arena footing. \\
    6 & Flip calf & Ties three of the calf’s legs with a piggin' string in a swift, practiced knot. \\
    7 & Tie legs & Throws hands up to signal completion and steps away as the calf remains bound for six seconds. \\
    8 & Raise hands & -- \\
    9 & Step back & -- \\
    10 & Stop time & -- \\

    \bottomrule
  \end{tabular}
  \caption{\textbf{Comparison of short-fixed and context-rich sub-action scripts for \textit{tie-down roping} in Rodeo.} The DTW alignment paths for both instances are shown in Figure~\ref{fig:SF-and-CR-comparison}}
  \label{tab:SF-and-CR-comparison}
\end{table*}

\section{VLM Evaluation Settings}
We provide here the detailed evaluation settings for the video–language models reported in Table~\ref{tab:zeroshot_baselines} of the main paper.

Our evaluation \textbf{strictly follows the official ActionAtlas\cite{salehi2024actionatlas} protocol and public leaderboard}. Each model is evaluated using \textbf{multiple-choice question–answer pairs}, where the model must select from a fixed set of candidate actions. As part of the standardized pre-processing pipeline, \textbf{all audio signals, speech transcripts, and any textual cues visible in the video are discarded or blurred} prior to inference. This ensures that evaluation reflects purely visual understanding rather than reliance on auxiliary textual or audio information.

Additionally, ActionAtlas provides a comprehensive analysis of how performance varies with different frame rates. The results below correspond to the \textbf{best-performing frame-rate configuration for each model} as reported in the official leaderboard. The number of frames, token configuration, and Top-1 accuracy for each model are shown below:

\begin{table}[h]
\centering
\begin{tabular}{lccc}
\hline
\textbf{Model} & \textbf{Frames} & \textbf{Tokens} & \textbf{Top-1 Acc. (\%)} \\
\hline
Qwen2-VL-7B & 16 & $8 \times 576$ & 30.24 \\
VideoLLaMA & 16 & $16 \times 256$ & 22.71 \\
VideoChat2 & 64 & $64 \times 196$ & 21.27 \\
mPLUG-Owl-Video & 16 & $16 \times 256$ & 19.49 \\
LLaVA-Next-Video-7B & 64 & $64 \times 144$ & 22.90 \\
\hline
\end{tabular}
\caption{Evaluation settings and results for baseline VLMs on the ActionAtlas benchmark.}
\label{tab:vlm_settings}
\end{table}

This table reproduces the official ActionAtlas leaderboard conditions under which we report all baseline results in the main text.

\end{document}